\documentclass[journal]{IEEEtran}

\usepackage{hyperref}
\usepackage{cite}
\usepackage{amssymb}
\usepackage{mathtools}
\usepackage{amsmath}
\usepackage[pdftex]{graphicx}
\usepackage{multirow}
\usepackage[caption=false,font=normalsize,labelfont=sf,textfont=sf]{subfig}
\usepackage{xcolor}

\begin{document}

\title{DAugNet: Unsupervised, Multi-source, Multi-target, and Life-long Domain Adaptation for Semantic Segmentation of Satellite Images}

\author{Onur Tasar,~\IEEEmembership{Student member,~IEEE,}
	    Alain Giros,\\
		Yuliya Tarabalka,~\IEEEmembership{Senior member,~IEEE,} 
		Pierre Alliez, 
		and S{\'e}bastien Clerc

\thanks{O.~Tasar, and P.~Alliez are with Universit{\'e} C{\^o}te d'Azur and Inria, Titane project-team, 06902 Sophia Antipolis, France. (E-mail: onur.tasar@inria.fr).}
\thanks{A.~Giros is with Centre National d'{\'E}tudes Spatiales, 18 Avenue Edouard Belin, 31400 Toulouse, France.}
\thanks{Y.~Tarabalka is with LuxCarta Technology, Parc d'Activit{\'e} l'Argile, Lot 119b, Mouans Sartoux 06370, France.}
\thanks{S.~Clerc is with ACRI-ST, 260 Route du Pin Montard, 06410 Biot, France.}}

\markboth{}%
{Shell \MakeLowercase{\textit{et al.}}: Bare Demo of IEEEtran.cls for IEEE Journals}

\maketitle

\begin{abstract}
The domain adaptation of satellite images has recently gained an increasing attention to overcome the limited generalization abilities of machine learning models when segmenting large-scale satellite images. Most of the existing approaches seek for adapting the model from one domain to another. However, such single-source and single-target setting prevents the methods from being scalable solutions, since nowadays multiple source and target domains having different data distributions are usually available. Besides, the continuous proliferation of satellite images necessitates the classifiers to adapt to continuously increasing data. We propose a novel approach, coined DAugNet, for unsupervised, multi-source, multi-target, and life-long domain adaptation of satellite images. It consists of a classifier and a data augmentor. The data augmentor, which is a shallow network, is able to perform style transfer between multiple satellite images in an unsupervised manner, even when new data are added over the time. In each training iteration, it provides the classifier with diversified data, which makes the classifier robust to large data distribution difference between the domains. Our extensive experiments prove that DAugNet significantly better generalizes to new geographic locations than the existing approaches. 
\end{abstract}


\IEEEpeerreviewmaketitle

\section{Introduction}~\label{sec:introduction}
\IEEEPARstart{W}{ith} the help of significant technological developments over the years, it has been possible to access to massive amounts of remote sensing data. For example, the constellations of Pl{\'e}iades satellites are able to capture large amounts of images with high spatial resolution over the globe. The availability of such enormous data has opened the door to various applications and raised many challenges. Among these challenges, semantic segmentation or dense labeling of satellite images has become one of the most interesting and long-standing problems, since it is a crucial process for a wide range of applications in domains such as urban monitoring, urban management, agriculture, automatic mapping, and navigation.

In the last decade, convolutional neural networks (CNNs) have been one of the most promising tools for semantic segmentation. Particularly for dense labeling of remote sensing data, especially U-net~\cite{ronneberger2015u} is a commonly used network architecture because of its great success in many benchmarks such as INRIA~\cite{maggiori2017can} and SpaceNet~\cite{van2018spacenet}. Although U-net and some other CNNs generate excellent maps when training data well represent test data, their performance significantly drops when there exists a large distribution shift between the training and the test data. In the literature, this issue is referred to as domain adaptation~\cite{tuia2016domain}. In the field of remote sensing, the domain shift usually stems from various atmospheric conditions and some differences in acquisition that change spectral characteristics of objects, intra-class variability, differences in the spectral characteristics of the sensor, or images with different types of spectral bands (e.g., some image might have red, green, and blue bands, whereas the others might consist of near infrared, red, and green channels.).

In remote sensing, we can refer to each satellite image as a domain. In a typical domain adaptation problem, the training and  the test images are usually designated as source and target domains. The most naive approach to hinder the performance deficiency under a large distribution difference between the domains is to manually annotate some portion of the target domain to fine-tune~\cite{maggiori2016convolutional} the already trained model on the source domain. However, such strategy requires human intervention in the process, and labeling even some portion of a satellite image is time-consuming. To increase the generalization capabilities of CNNs, another common approach is to perform online data augmentation with random color change~\cite{buslaev2020albumentations}. For instance, gamma correction and random contrast change are widespread used in remote sensing~\cite{tasar2019incremental}. However, these augmentation approaches are limited when it comes to adapting the model from one domain to another, when there exists a significant difference between data distributions. For example, with this simple augmentation approaches, it is not possible to adapt a model from one domain containing red, green, blue bands to another one composing of near-infrared, red, green channels. To overcome this limitation, one can use generative adversarial networks (GANs)~\cite{goodfellow2014generative} to generate a fake source domain with a similar data distribution to that of target domain. The fake data then can be used to train a classifier.

Unsupervised domain adaptation aims at adapting the classifier to unlabeled target domains. The common single-source and single-target adaptation setting limits the proposed approaches to small-scale segmentation problems. The recent generation of satellites with a short revisit time generate in routine massive amounts of remote sensing data. These data usually contain multiple domains with largely different data distributions and are relevant for many real-world applications. Therefore, to perform large-scale classification, an ideal unsupervised domain adaptation approach should be capable of learning from annotated multiple source domains and well segmenting unlabeled multiple target domains even when the data distributions of all the domains are different. In addition, nowadays, the constellations of satellites collect images from the entire earth everyday. As a consequence, in many cases, one receives new annotated source domains and unlabeled target domains over the time. Hence, it is also crucial for the ideal approach to perform life-long adaptation over the continuously increasing data.

By performing unsupervised style transfer between two domains with GANs, it is possible to generate a fake source domain whose data distribution is as similar as possible to the distribution of a target domain~\cite{tasar2020colormapgan, tasar2020semi2i}. A possible solution for unsupervised, multi-source, multi target, and life-long domain adaptation problem would be learning to perform style transfer between every domain pairs. If this task can be accomplished, we can generate multiple fake domains from each domain, where each fake domain is representative for a different domain. We then can use the fake domains generated from the source domains to train a classifier. Training a classifier on such diversified data would allow the classifier to learn from the data that are representative for both the source and the target domains, instead of capturing the idiosyncrasies of each source domain.

However, the problem of style transfer between progressively growing multiple domains raises some challenges. Firstly, all of the fake domains generated from each source domain must be semantically consistent with the original domain. For example, if the method replaces some objects with others or adds artificial structures to the fake domains in the generation process, the fake domains and the ground-truth for the real domain would not match. Thus, we cannot train a classifier. Secondly, when we perform style transfer between domain pairs, the data distribution of the fake domain generated from one domain must be similar to the distribution of the other domain. Otherwise, the fake domain would not be representative for the other domain. Thirdly, the number of networks should not grow with the number of domains to scale the method to many domains. For example, if the method requires using a style encoder for each domain as in MUNIT~\cite{huang2018multimodal}, DRIT~\cite{lee2018diverse}, StarGAN~\cite{choi2018stargan}, StarGAN~v2~\cite{choi2019stargan}, and StandardGAN~\cite{tasar2020standardgan}, it would be limited to a certain number of domains. In addition, it would be necessary to add new style encoders every time when we receive new data. Finally, once the training is completed, ability to perform style transfer between the domains on the fly is desired. Otherwise, it would be required to store all the fake domains on a disk.

In this article, we introduce a new approach dealing with the aforementioned challenges. Our contributions are as follows:
\begin{enumerate}
	\item We propose an approach that can perform style transfer between multiple domains with only one encoder, one decoder, and one discriminator unlike MUNIT~\cite{huang2018multimodal}, DRIT~\cite{lee2018diverse}, StarGAN~\cite{choi2018stargan}, StarGAN v2~\cite{choi2019stargan}, and StandardGAN~\cite{tasar2020standardgan} that rely upon a different style encoder for each domain. In addition, since the number of networks is constant irrespective of the number of domains, our approach can do style transfer across multiple domains even when new domains are progressively added. 
	\item We introduce novel DAugNet consisting of a data augmentor and a classifier. The data augmentor, which is a shallow network, can stylize each domain as another one. Due to its simple architecture, it allows online data diversification rather than storing all the fake data on a disk and loading them when training a classifier. In each training iteration of DAugNet, the data augmentor diversifies a batch of training patches by stylizing each patch as a randomly selected domain before passing the batch to the classifier.  
	\item We validate DAugNet on semantic segmentation of Pl{\'e}iades images captured over multiple cities from three European countries. Our data comprise images with either red, green, blue bands or near-infrared, red, green bands. In three extensive experiments, we investigate how DAugNet performs on single-source and single-target, multi-source and multi-target, and life-long adaptation problems. We also conduct five ablation studies to further analyze the properties of our solution.
\end{enumerate}

\section{Related Work}
\paragraph{Adapting the Inputs} These methods usually aim at modifying the source domain to make its data distribution similar to the distribution of the target domain. Then, a classifier is trained on the modified source domain and evaluated on the original target domain. Histogram matching~\cite{Gonzalez, inamdar2008multidimensional} and graph matching~\cite{tuia2012graph} are commonly used methods for the above-mentioned modification. Another way to adapt the source domain to the target domain is to use style transfer or image-to-image translation (I2I) methods to generate a fake source domain stylized as the target domain. For instance, CyCADA~\cite{hoffman2017cycada} generates a fake source domain with CycleGAN~\cite{zhu2017unpaired} and uses it to train a classifier. Similarly, Benjdira~\textit{et~al.}~\cite{benjdira2019unsupervised} use  CycleGAN for I2I between aerial images. Tasar~\textit{et~al.} have recently proposed ColorMapGAN~\cite{tasar2020colormapgan} and SemI2I\cite{tasar2020semi2i} to perform domain adaption between satellite image pairs. Other state-of-the-art I2I approaches like UNIT~\cite{liu2017unsupervised}, MUNIT~\cite{huang2018multimodal}, DRIT~\cite{lee2018diverse} are also relevant for domain adaptation.

\paragraph{Adapting the classifiers} The goal of these methods is to adapt the classifier to an unlabeled target domain rather than modifying the inputs. To achieve this task, a common approach is simultaneously to learn from the source domain supervisedly and to align the features extracted from both domains with adversarial learning~\cite{hoffman2016fcns, tsai2018learning, huang2018domain}. The same strategy has also been applied to satellite images~\cite{deng2019large}. To adapt a classifier to an unlabeled domain, the approaches based on self learning~\cite{zhang2018fully, zou2018unsupervised}, expectation-maximization (EM)~\cite{bruzzone2001unsupervised, bruzzone2002partially}, multiple cascade classifiers~\cite{bruzzone2002multiple}, variants of support vector machine (SVM)~\cite{bruzzone2009domain}, centroid and covariance alignment~\cite{ma2018centroid}, and tensor alignment~\cite{qin2019tensor} have been proposed as well.

\paragraph{Data Standardization} The objective of the data standardization methods is to standardize each domain in the pre-processing step so that all the domains have similar data distributions. After the standardization, one can work only on the standardized data instead of the real data. One of the most widespread used standardization method is referred to as linear scaling to unit variance (a.k.a. z-score normalization)~\cite{aksoy2001feature}, which is computed by subtracting the mean value from the features and dividing by the standard deviation. Another common approach is histogram equalization~\cite{Gonzalez}. The color constancy algorithms~\cite{agarwal2006overview} such as gray-world~\cite{buchsbaum1980spatial} and gamut~\cite{forsyth1990novel} aim at extracting the intrinsic properties on the objects that are affected by the illuminant. These algorithms can also be used for data standardization. Tasar~\textit{et~al.} have recently proposed StandardGAN~\cite{tasar2020standardgan} with the purpose of standardizing multiple domains with GANs. However, the approach requires using a style encoder for each domain, which limits the method to a certain number of domains. In some cases, the domain shift can be corrected by radiometric correction~\cite{pacifici2014importance, itten1993geometric} as well. Note that the data standardization methods can be used for multi-source and multi-target adaptation problems.

\paragraph{Multi-source Adaptation} One can consider doing I2I between each source and target domain pairs to make the data distribution of the source domains similar to the distribution of the target domain. However, the necessity of applying an I2I approach many times makes this strategy inconvenient. Zhao~\textit{et~al.}~\cite{zhao2019multi} extends CyCADA~\cite{hoffman2017cycada} to multi-source single-target domain adaptation problem. However, due to its extensive computational requirements, it is not possible to train the method end-to-end on the current GPUs. The authors split the training pipeline into several stages, where the stages are executed consecutively and outputs of each stage are passed to the next stage as inputs. Redko~\textit{et~al.} introduce an approach based on optimal transport~\cite{redko2018optimal}. Elshamli~\textit{et~al.} propose a new method for local climate zone classification. However, the method may not be appropriate for less coarse classes such as building and road. Several approaches have been proposed for multi-source domain adaption in the context of classification~\cite{zhao2018adversarial, xu2018deep, peng2019moment}. However, these methods are not appropriate for semantic segmentation task, as it requires generating a precisely structured pixel-wise classification.

\paragraph{Multi-target Adaptation}
As in multi-source adaptation problem, it is possible to solve multi-target adaptation problem by executing an I2I method to generate fake source domains with similar distributions to those of the target domains. These fake domains can then be used to train a classifier. However, such solution is sub-optimal as it is needed to apply an I2I approach multiple times. Specifically for multi-target adaptation problem, an information theoretic approach~\cite{gholami2020unsupervised} and a model parameter adaptation framework~\cite{yu2018multi} have been proposed. However, in their problem definitions, these methods assume that there exist only one source domain. In our case, on the other hand, there might be more than one source domain. In addition, these methods have been proposed for image categorization problem. Hence, it may not be possible to use them for semantic segmentation.

\paragraph{Life-long Adaptation}
The problem of life-long, continual, incremental, or gradual adaptation has been tackled via GANs~\cite{wulfmeier2018incremental}, meta-learning~\cite{zhang2019online}, or self training~\cite{kumar2020understanding}. The main limitation of these works is their assumption that progressively added domains are a series of intermediate domains. Although these method are suitable for problems such as video processing and autonomous driving, they may not be applicable to remote sensing data, as the incoming domains might be completely different from each other. In the remote sensing literature, there are methods based on semi-supervised active learning~\cite{persello2012active, matasci2012svm, deng2018active, ghassemi2019learning}, where an annotator labels some portion of the received domain to be included in the training data. However, we seek for an unsupervised method instead, as automation is the key for practical real-world applications.

\begin{figure}
\centering
    \includegraphics[width=\linewidth]{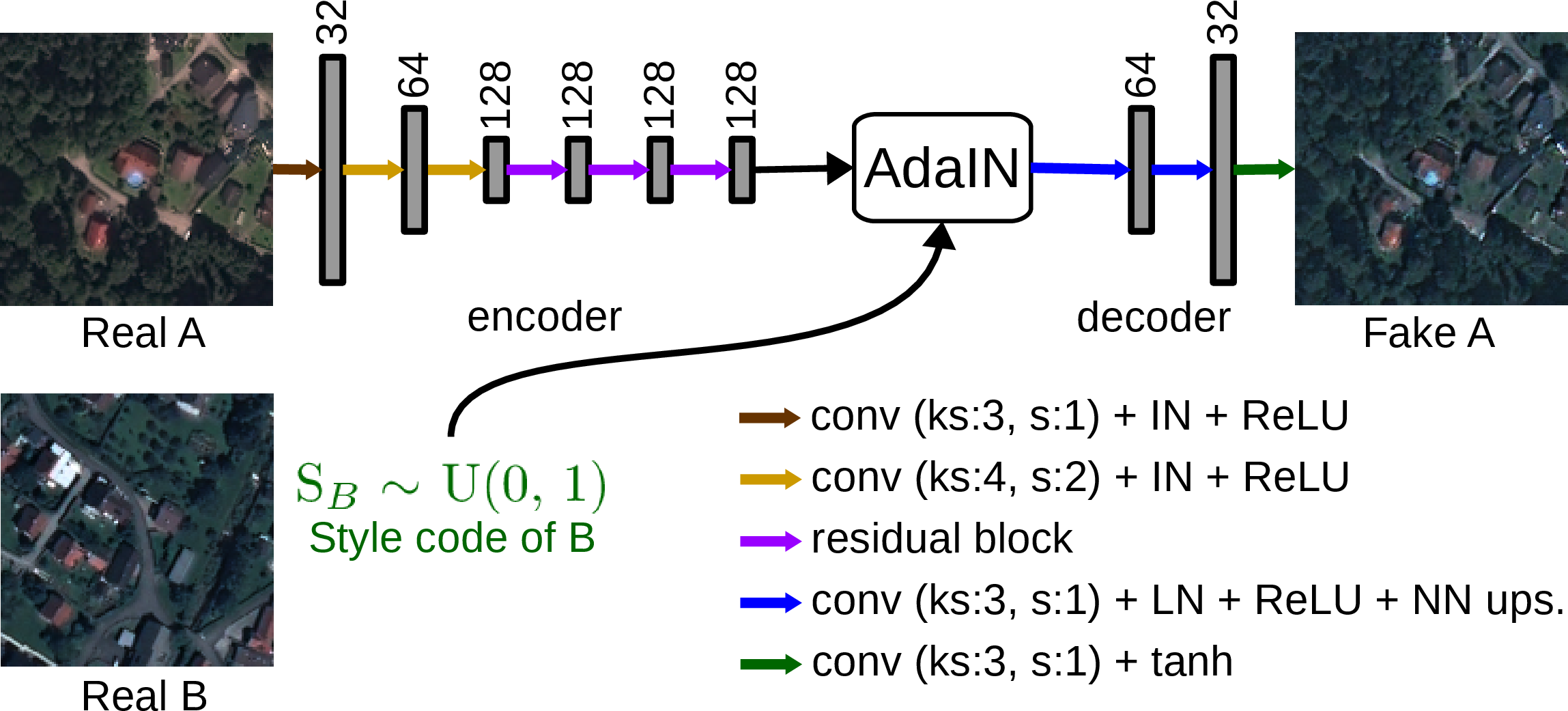}
\caption{Combining the content of one domain with the style of another domain. $ks$ and $s$ correspond to kernel size and stride parameters of the convolution. $IN$, $LN$, and $NN~ups.$  stand for instance normalization, layer normalization, and the nearest neighbor upsampling, respectively. The numbers above the gray rectangles denote the number of channels in each activation.}
\label{fig:adaIN}
\end{figure}

\begin{figure}
\centering
    \includegraphics[width=\linewidth]{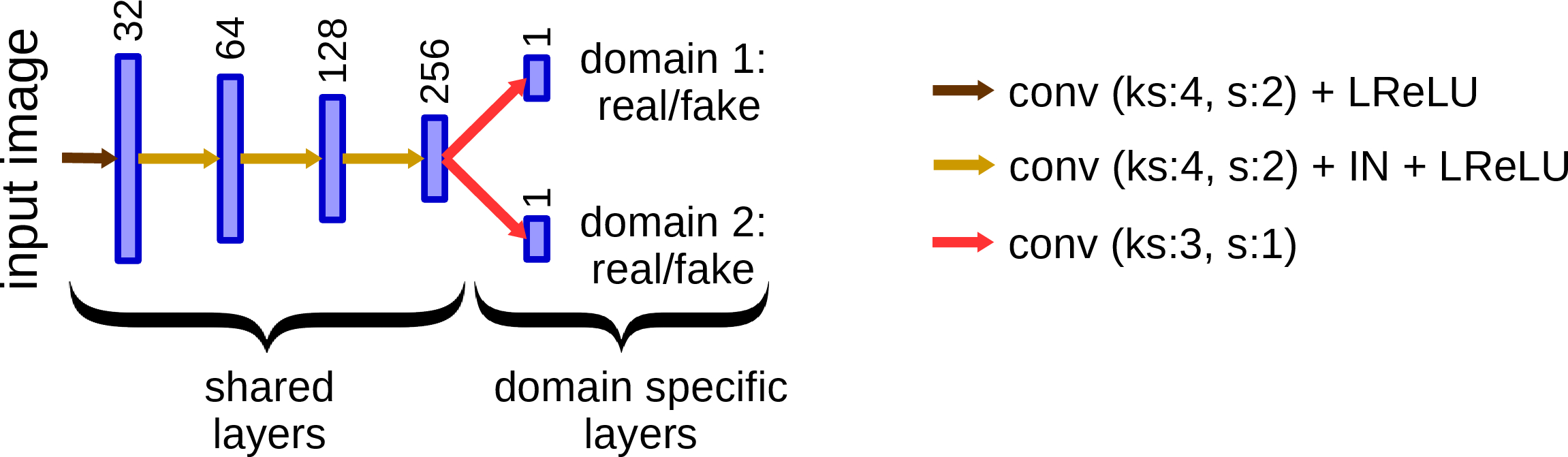}
\caption{The multi-task discriminator. $IN$ and $LReLU$ denote instance normalization and leaky rectified linear unit, respectively. The number of channels in each activation is indicated above the blue rectangles.}
\label{fig:discriminator}
\end{figure}

\begin{figure*}
\centering
    \includegraphics[width=\linewidth]{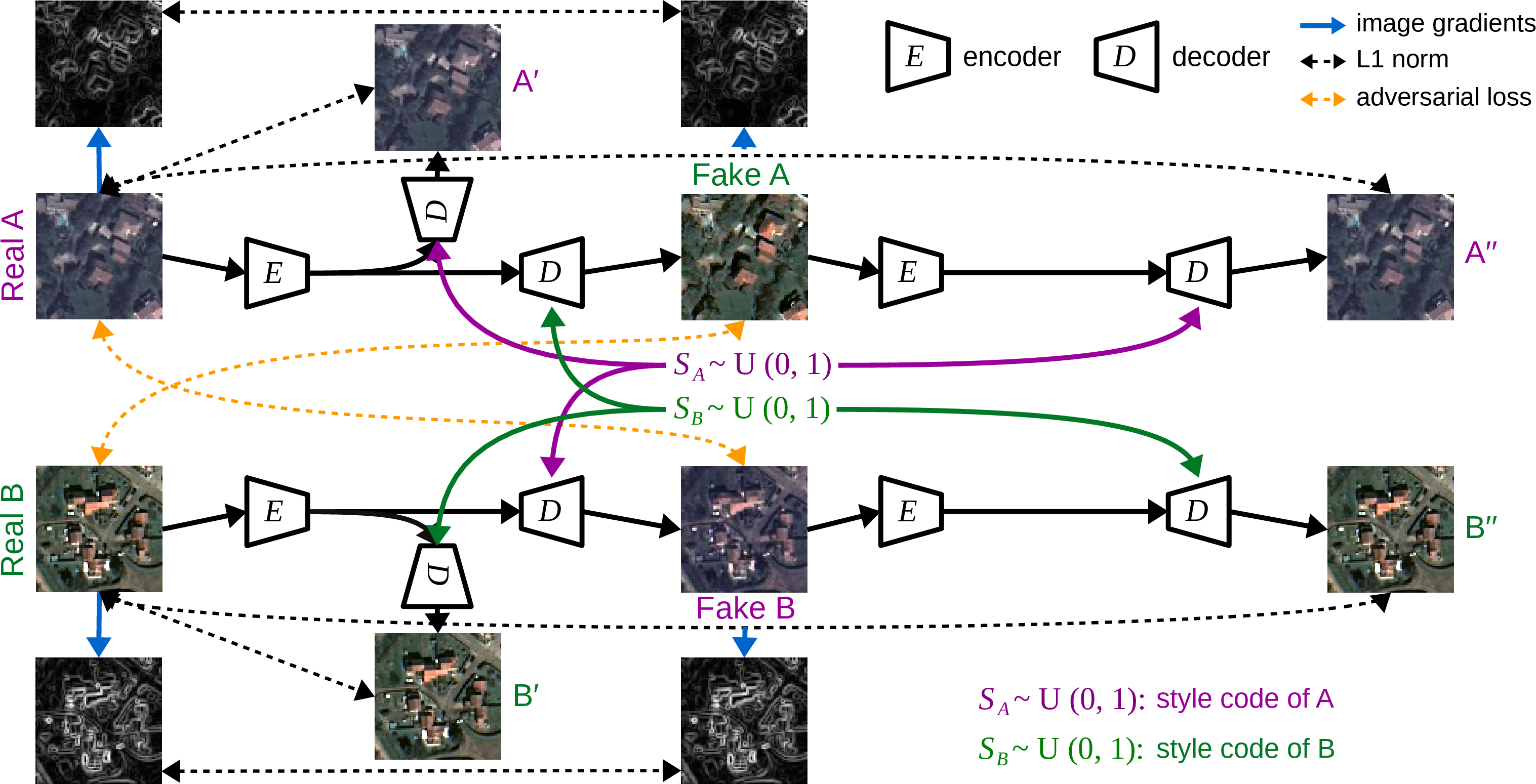}
\caption{Style transfer between two domains. Before the training starts unique and style random codes $S_A$ and $S_B$ are randomly drawn from the uniform distribution. To generate fake~A and fake~B, we combine the embeddings of A and B with $S_B$ and $S_A$, respectively. L1 norms enforce semantic consistency between A and fake A, and B and fake B. The adversarial losses force fake A and fake B to look like B and A, respectively.}
\label{fig:I2I}
\end{figure*}

\section{Method}
Our training pipeline consists of two stages. In the first stage, we learn how to stylize each domain like any other using only one encoder, one decoder, and one discriminator. Note that this stage is completely unsupervised. Therefore, we use both annotated source and unlabeled target domains during training. A subset of the networks used in this stage forms a data augmentor. In the second stage, we train DAugNet that is composed of the data augmentor and a classifier. In each training iteration of DAugNet, the data augmentor is frozen and diversifies a batch of patches sampled from source domains by stylizing each patch in the batch as a randomly selected domain. The diversified batch is then given to the classifier. The randomly selected domain can be chosen among both source and target domains. Hence, when training DAugNet, the classifier learns from the data that are representative for all of the source and the target domains. As a result, it is expected that DAugNet would perform better than a classifier trained only on real source domains.

In the rest of this section, we first describe how we concurrently switch the styles of two domains. We then detail how we perform style transfer between multiple domains. We finally explain the architecture of DAugNet and its overall training pipeline for multi-source, multi-target, and life-long domain adaptation.

\subsection{Style Transfer Between Two Domains}\label{sec:i2i}
To perform style transfer between two domains, we use one encoder, one decoder, and one multi-task discriminator. Throughout this sub-section, we refer to one of the domains as A and the other one as B for the sake of simplicity. 

Before the training starts, we initially generate a unique and constant style code for each domain. We denote by $S_A = \{ \gamma_A, \beta_A \}$ and $S_B = \{ \gamma_B, \beta_B \}$ the style codes of A and B. We initialize each parameter of $S_A$ and $S_B$ (i.e., $\gamma$ and $\beta$) with 128 values drawn from the uniform distribution that ranges between 0 and 1. To generate a fake domain stylized as the other one, we first extract an embedding from the domain using the encoder. We then normalize the embedding via adaptive instance normalization (AdaIN)~\cite{huang2017arbitrary} that is defined as:
\begin{equation}\label{eq:adaIN}
\text{AdaIN}( x, \gamma, \beta ) = \gamma \left( \frac{x - \mu(x)}{\sigma(x)} \right) + \beta,
\end{equation}
where x denotes the final activation of the encoder from one domain, and $\gamma$ and $\beta$ correspond to the style code parameters of the other domain. We finally decode the normalized embedding via the decoder. As depicted in Fig.~\ref{eq:adaIN}, the embedding extracted from one domain through the encoder consists of 128 channels, each of which is scaled and shifted by the style code parameters of the other domain with Eq.~\ref{eq:adaIN} before feeding the embedding to the decoder. Decoding the normalized embedding of A with $S_B$ and the embedding of B with $S_A$ results in generating fake A stylized as B and fake B with the style of A. StandardGAN~\cite{tasar2020standardgan} aims at learning the parameters of $S_A$ and $S_B$ from A and B by domain specific so-called style encoders. Such approach requires using a different style encoder for each domain. However, we have discovered that it is possible to combine the content of A with the style B (or vice versa) using randomly initialized and unique $S_A$ and $S_B$ parameters instead of trying to learn them. 

After combining A with $S_B$ and B with $S_A$ to generate fake A and fake B, we need to ensure that the data distributions of A and fake B, and B and fake A are as similar as possible. To overcome this issue, one could use adversarial learning to force the distributions of fake A and fake B to be similar to those of B and A, respectively. To perform style transfer between two domains, the current state-of-the-art I2I approaches like SemI2I~\cite{tasar2020semi2i}, CycleGAN~\cite{zhu2017unpaired}, UNIT~\cite{liu2017unsupervised} use two discriminators, where one discriminates between A and fake B, and the other one tries to distinguish B from fake A. However, as discussed in Sec.~\ref{sec:introduction}, one of the challenges is to use as small number of networks as possible to easily extend the method to multi-domain adaptation problem. Hence, instead of including multiple discriminators, we propose to use only one multi-task discriminator. As shown in Fig.~\ref{fig:discriminator}, our discriminator comprises several shared layers and multiple domain specific layers. Each domain specific layer tries to understand whether the given data are fake or belong to that domain. Let us also remark that each domain specific layer of our discriminator outputs a two dimensional matrix. Each element of this matrix determines whether different parts of the input image are real or fake. We then take average of all the elements to generate a scalar quantifying the realness of the input. Our generator is depicted in Fig.~\ref{fig:I2I}. Let us denote by $D_A$ and $D_B$ the discriminator's outputs for A and B, and by $G$ the generator. The adversarial loss for the discriminator is defined as:
\begin{equation}
\mathcal{L}_{D_{adv}}{(X, Y)} = -\big[ \mathbb{E}_{x}[\text{log}(D_{X}(x))] +  \mathbb{E}_{y}[\text{log}(1 - D_{X}(G(y)))] \big],
\end{equation}
where $X$ and $Y$ denote two domains, and $x$ and $y$ correspond to patches sampled from these domains. The adversarial loss for the generator is described as:
\begin{equation}
\mathcal{L}_{G_{adv}}{(X, Y)} = -\big[ \text{log} (D_{X}(G(y))) \big].
\end{equation}

We compute the overall adversarial losses for the discriminator and the generator as:
\begin{equation}
\mathcal{L}_{D_{adv}} = \mathcal{L}_{D_{adv}}(\text{A, B}) + \mathcal{L}_{D_{adv}}(\text{B, A})
\end{equation}
and
\begin{equation}
\mathcal{L}_{G_{adv}} = \mathcal{L}_{G_{adv}}(\text{A, B}) + \mathcal{L}_{G_{adv}}(\text{B, A}).
\end{equation}

Another challenge is to keep A and fake A, and B and fake B semantically consistent. Otherwise fake A and fake B would not match with the ground-truth of A and B, and could not be used to train a classifier. To enforce the semantic consistency, we define several constraints. Firstly, after we generate fake A and fake B, we combine their contents with the style codes of B and A (i.e., $S_B$ and $S_A$) to generate A$''$ and B$''$. By switching the styles of fake A and fake B, the original domains need to be reconstructed. Hence, we minimize the cross reconstruction loss $\mathcal{L}_{cross}$ that is defined as:
\begin{equation}
\mathcal{L}_{cross} = \lvert \text{A} - \text{A}'' \rvert + \lvert \text{B} - \text{B}'' \rvert.
\end{equation}
Secondly, when we combine A with $S_A$ and B with $S_B$, we obtain A$'$ and B$'$. Here, since we combine the content of each domain with its own style, A$'$ and B$'$ must be the same as A and B. Therefore, we minimize the self reconstruction loss that is computed as:
\begin{equation}
\mathcal{L}_{self} = \lvert \text{A} - \text{A}' \rvert + \lvert \text{B} - \text{B}' \rvert.
\end{equation}
Finally, the textural features of A and fake A, and B and fake B must be very close. Let us assume that $Gr(\cdot, \cdot)$ is a function that takes two three-band images as inputs, converts them to gray-scale images, computes their horizontal and vertical gradients by Sobel filter, and sums L1 norm between the horizontal and L1 norm between the vertical gradients. We minimize the edge loss $\mathcal{L}_{edge}$ as:
\begin{equation}
\mathcal{L}_{edge} = Gr(\text{A}, \text{fake A}) + Gr(\text{B}, \text{fake B}).
\end{equation}
Note that other textural features such as Haralick features~\cite{haralick1973textural} can be considered as well. However, we prefer Sobel operator mainly because of its short execution time.

The final objective for the generator is computed as:
\begin{equation}\label{eq:g_loss}
\mathcal{L}_{G} = \lambda_{1} \mathcal{L}_{adv\_G} + \lambda_{2} \mathcal{L}_{cross} + \lambda_{3} \mathcal{L}_{self} + \lambda_{4} \mathcal{L}_{edge}, 
\end{equation}
where $\lambda_{1}, \lambda_{2}, \lambda_{3}$, and $\lambda_{4}$ adjust the relative importance of each loss. The discriminator loss is defined as:
\begin{equation}\label{eq:d_loss}
\mathcal{L}_{D} = \lambda_{1} \mathcal{L}_{adv\_D}.
\end{equation}
In the training stage, we minimize both $\mathcal{L}_{G}$ and $\mathcal{L}_{D}$ simultaneously.
\begin{figure}
\centering
    \includegraphics[width=\linewidth]{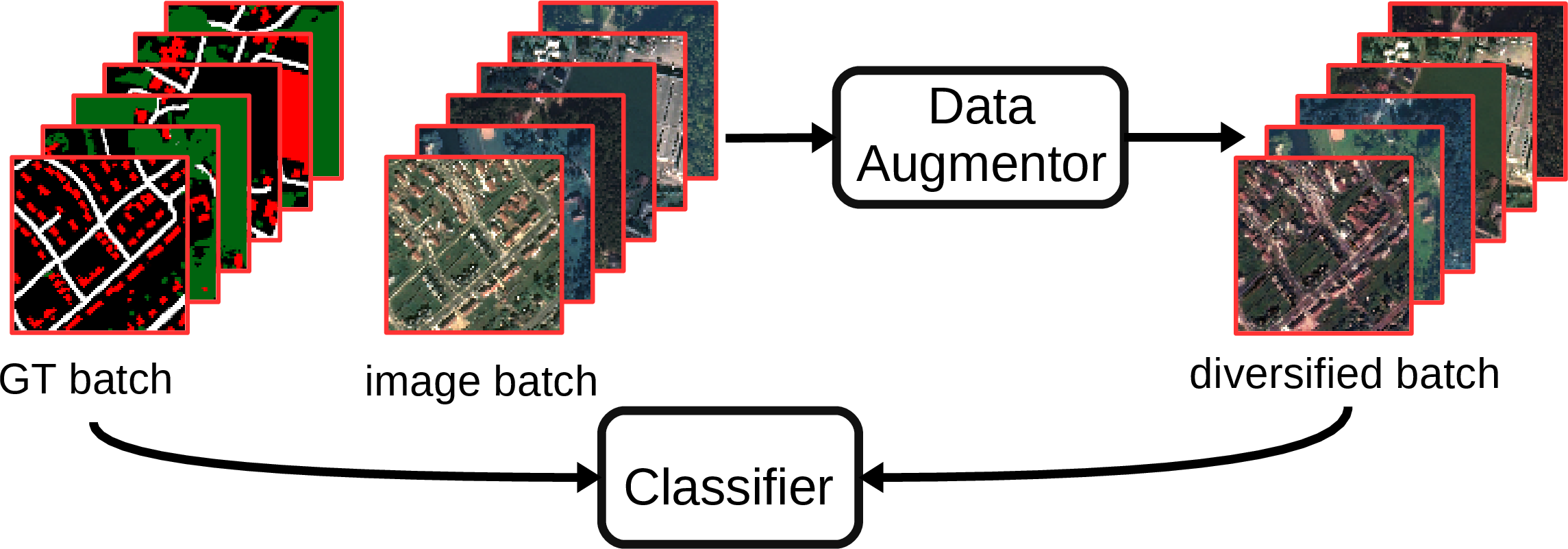}
\caption{The training procedure of DAugNet that comprises a data augmentor and a classifier. In each training iteration, the classifier learns from the diversified batch generated by the data augmentor.}
\label{fig:DAugNet}
\end{figure}
\subsection{DAugNet for Multi-source, Multi-Target, and Life-long Domain Adaptation}~\label{sec:daugnet}
Let us assume that we have $N$ domains out of which $N_{s}$ of them are annotated source domains and $N_t$ of them are unlabeled target domains. We also suppose that all of the domains have considerably different data distributions. In such a multi-source and multi-target segmentation setting, our goal is to generate maps for $N_{t}$ unlabeled target domains. In the pre-processing step, we first split all of the domains into smaller patches and generate a unique style code for each domain. We then perform multi-domain style transfer by sampling a patch from randomly selected two domains among $N$ and minimizing Eqs.~\ref{eq:g_loss} and \ref{eq:d_loss} in each training iteration. After a sufficiently long training process, we are able to generate fake $N-1$ domains from each one. Our constraints explained in Sec.~\ref{sec:i2i} enforce each fake domain to representative for a different domain and to be semantically consistent with the original domain.

Once the training process of the initial unsupervised and multi-domain style transfer is completed, we move on to training DAugNet. As illustrated in Fig.~\ref{fig:DAugNet}, the main components of DAugNet is the data augmentor and the classifier. In this stage, the data augmentor is frozen, only the classifier is updated. The combination of the encoder and the decoder forms a data augmentor. Both the encoder and the decoder are shallow networks, where the former consists of six layers and the latter is composed of only two layers (see Fig.~\ref{fig:adaIN}). In each training iteration of DAugNet, we randomly sample a batch of training patches from $N_s$ source domains as well as their corresponding ground-truth. With 0.9 probability, the data augmentor diversifies the batch by extracting its embedding via the encoder, scaling and shifting the embedding for each patch with the style code parameters of a randomly selected domain, and decoding the normalized embedding. We then use the diversified batch and the ground-truth for the original batch to update the classifier by minimizing the classification loss $\mathcal{L}_{class}$ defined as:
\begin{equation}\label{eq:class_loss}
\mathcal{L}_{class} = \lambda_5 \mathcal{L}_{sigmCE}(y, \hat{y}) + \lambda_6 \mathcal{L}_{soft\_IoU}(y, \hat{y}),
\end{equation}
where $\mathcal{L}_{sigmCE}$, $\mathcal{L}_{soft\_IoU}$, $y$, and $\hat{y}$ denote sigmoid cross entropy loss, soft IoU loss~\cite{mattyus2017deeproadmapper}, class labels, and the predictions, respectively. Diversifying the original batch with the proposed shallow data augmentor has two advantages. Firstly, such diversification approach allows the classifier to learn from the data that are representative for all $N$ domains. Therefore, for target domains, we can expect DAugNet to achieve a better performance than naively training a classifier on the real data. Secondly, the diversification can be carried out in an online fashion, rather than storing all the fake domains in the pre-processing step and loading them when training the classifier. In each iteration, we deactivate the data augmentor with 10$\%$ chance for the classifier to learn from the real data as well. 

To better explain life-long, multi-source, and multi-task adaptation problem, let's assume that we are provided additional $M$ domains composing of $M_s$ annotated source and $M_t$ unlabeled target domains with different data distributions. The problem now becomes segmenting $N_t + M_t$ domains. To solve this problem, we first generate unique  style codes for newly added $M$ domains. We load the style codes of the initial $N$ domains and the pre-trained weights for the encoder, the decoder, and the classifier of DAugNet. We also initialize the parameters of the discriminator's shared layers as well as its domain specific layers for the previous domains with the pre-trained weights. For the current $M$ domains, we add additional $M$ domain specific layers to the discriminator with random initialization. We then conduct the initial training to perform style transfer between $N + M$ domains. This time, when selecting two random domains, we randomly choose one domain among the current $M$ domains and another one from all $M + N$ domains we have at hand. With this sampling strategy, we guarantee that at least one of the domains is selected from the newly added ones. We finally fine-tune DAugNet with the same diversification method as explained above. Let us remark that every time when we receive a new domain, we do not increase the number of networks. We only add a new layer to the discriminator. Considering that the networks with hundreds of layers can be fit into the current GPUs as in ResNet-152~\cite{he2016deep}, it is feasible to add many layers to the discriminator.

\section{Experiments}
\subsection{Data Set}
\begin{table}
\centering
\caption{The data set.}
\label{table:data_stats}
\begin{tabular}{l|c|c|c|c}
\hline
\multirow{2}{*}{\textbf{City (Country)}} & \multicolumn{3}{c|}{\textbf{Class percentages ($\%$)}} & \textbf{256$\times$256} \\ 
\cline{2-4}
                                         & \textbf{Building} & \textbf{Road} & \textbf{Tree}      & \textbf{patches}  \\
\hline
Villach (AT)                             &  9.26             & 10.63         & 19.91              &  749     \\
Sankt P{\"o}lten (AT)                    &  6.68             &  6.39         & 25.13              & 1558     \\
Bad Ischl (AT)                           &  5.51             &  6.0          & 35.38              &  457     \\
Salzburg Stadt (AT)                      &  9.44             &  8.69         & 23.88              & 2496     \\
Leibnitz (AT)                            &  7.00             &  7.37         & 16.78              &  572     \\
Bourges (FR)                             &  9.81             & 10.52         & 14.83              & 1188     \\
Lille (FR)                               & 18.36             & 12.71         & 15.40              & 2181     \\
B{\'e}ziers (FR)                         & 19.09             & 17.62         & 10.91              &  407     \\
Albi (FR)                                & 17.20             & 14.48         & 15.19              &  413     \\
Vaduz (LI)                               &  3.57             &  4.30         & 33.69              & 1612     \\
\hline
\end{tabular}
\end{table}
Our data set consists of Pl{\'e}iades images collected over five cities in Austria, four cities in France, and one city in Liechtenstein. We were given the annotations for building, road, and tree classes by the data providers.~\footnote{The authors would like to thank LuxCarta for providing us with the annotated data that allowed us to conduct this research.} Table~\ref{table:data_stats} reports city names and what percentage of the pixels belong to each class. The image from Leibnitz is composed of near-infrared, red, green bands, and its spatial resolution is 0.5~m. The rest of the cities comprise red, green, and blue channels, and their resolution is 1~m. We reduce the resolution of the image from Leibnitz to 1~m so as to make our data set homogenious in terms of spatial resolution. The images in our data set cover the total area of 663.02~km$^2$.

In the pre-processing step, we split all the images into 256$\times$256 smaller patches with an overlap of 32 pixels. We use these patches in every stage of our training pipeline (i.e., both the multi-domain style transfer part and the process of training DAugNet). We indicate the number of patches belonging to each city in Table~\ref{table:data_stats}.

\subsection{City-to-city Adaptation}
\begin{table*}
\centering
\caption{IoU scores for the first experiment (City-to-city adaptation).}
\label{table:iou_exp1}
\begin{tabular}{||p{0.01cm}p{0.1cm}|c||c|c|c|c|c||c|c|c|c||}
\hline			
\multicolumn{4}{||c|}{\textbf{Method}} & \multicolumn{4}{c||}{\textbf{Source: Bad Ischl $\Rightarrow$ Target: Villach}} & \multicolumn{4}{c||}{\textbf{Source: Villach $\Rightarrow$ Target: Bad Ischl}} \\
\hline                         
& \multicolumn{2}{c||}{\textbf{Name}} & \multicolumn{1}{c|}{\textbf{Type}} & \textbf{Building} & \textbf{Road} & \textbf{Tree} & \textbf{Overall} 
& \textbf{Building} & \textbf{Road} & \textbf{Tree} & \textbf{Overall} \\ 
\hline
\multicolumn{3}{||c||}{U-net~\cite{ronneberger2015u}}  & naive & 29.00 & 3.78 & 58.52 & 30.43 & 5.12 & 0.08 & 0.00 & 1.73 \\
\hline
\multicolumn{3}{||c||}{AdaptSegNet Single\cite{tsai2018learning}} & adapting &  6.01 &  4.37 & 10.43 &  6.94 &  3.06 &  2.71 & 10.23 &  5.33 \\
\multicolumn{3}{||c||}{AdaptSegNet Multi\cite{tsai2018learning}}  & classifier & 24.59 &  9.02 & 56.08 & 29.86 & 14.26 &  4.46 & 24.66 & 14.46 \\
\hline
\multirow{11}{*}{\rotatebox{90}{U-net trained on}}& \multirow{12}{*}{\rotatebox{90}{data generated by}}
  & Gray-world~\cite{buchsbaum1980spatial} & \multirow{2}{*}{data}  & 11.98 & 25.66 & 60.91 & 32.85 & 34.30 & 29.08 & 58.12 & 40.50 \\
& & Hist. Equalization~\cite{Gonzalez}     & \multirow{2}{*}{standardization}  & 23.28 & 23.36 & 64.31 & 36.98 & 34.08 & 22.81 & 64.93 & 40.60 \\
& & Z-score norm.~\cite{aksoy2001feature}  &   & 17.86 & 20.53 & 61.64 & 33.34 & 23.20 & 34.22 & 52.00 & 36.47 \\
\cline{3-12}
& & Hist. Matching~\cite{Gonzalez}         &  &  8.20 & 20.18 & 54.81 & 27.73 &  2.27 &  0.13 &  0.00 &  0.80 \\
& & UNIT~\cite{liu2017unsupervised}        & image &  1.96 &  0.51 & 66.68 & 23.05 & 32.51 &  4.39 & 41.77 & 26.22 \\
& & MUNIT~\cite{huang2018multimodal}       & to &  6.72 &  0.00 & 37.32 & 14.68 &  2.41 &  0.00 &  1.22 &  1.21 \\
& & DRIT~\cite{lee2018diverse}             & image &  0.00 &  0.00 & 15.14 &  5.05 &  0.00 &  0.30 &  0.00 &  0.10 \\
& & CycleGAN~\cite{zhu2017unpaired}        & translation & 25.80 & 25.20 & 66.07 & 39.03 & 42.01 & 12.15 & 76.39 & 43.52 \\
& & ColorMapGAN~\cite{tasar2020colormapgan}&  & 42.28 & \textbf{40.25} & 65.28 & 49.27 & 52.00 & \textbf{42.29} & 47.28 & 47.19 \\
& & SemI2I~\cite{tasar2020semi2i}          &  & 40.18 & 32.66 & 66.83 & 46.56 & 49.32 & 41.06 & 70.18 & 53.52 \\
\hline
\multicolumn{3}{||c||}{DAugNet (ours)}     & diversification & \textbf{49.68} & 35.32 & \textbf{68.14} & \textbf{51.05} & \textbf{53.19} & 41.84 & \textbf{80.07} & \textbf{58.37} \\
\hline  
\end{tabular}
\end{table*}

\begin{figure*}
\centering
\subfloat[]{\includegraphics[width=0.15\linewidth]{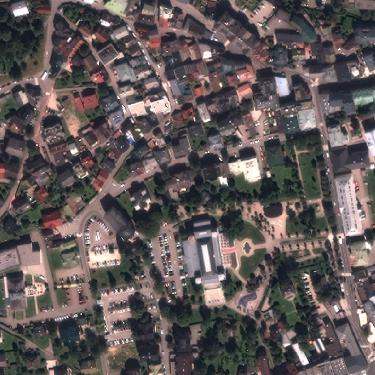}\label{fig:exp1_real_bad_ischl}}
\hfill
\subfloat[]{\includegraphics[width=0.15\linewidth]{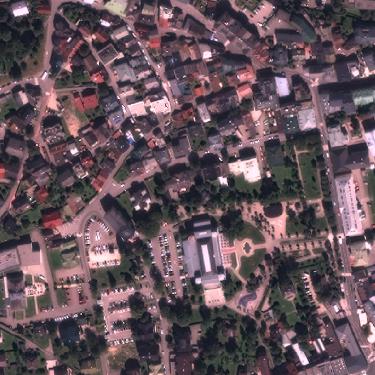}}
\hfill
\subfloat[]{\includegraphics[width=0.15\linewidth]{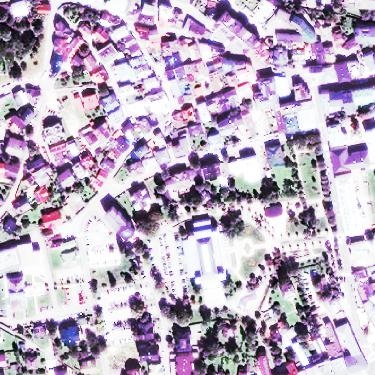}}
\hfill
\subfloat[]{\includegraphics[width=0.15\linewidth]{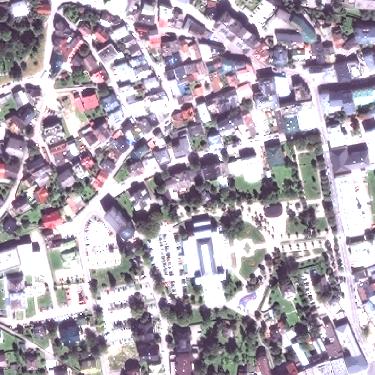}}
\hfill
\subfloat[]{\includegraphics[width=0.15\linewidth]{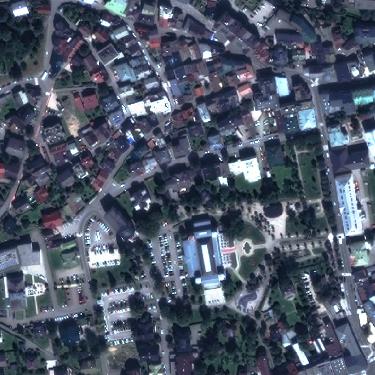}}
\hfill
\subfloat[]{\includegraphics[width=0.15\linewidth]{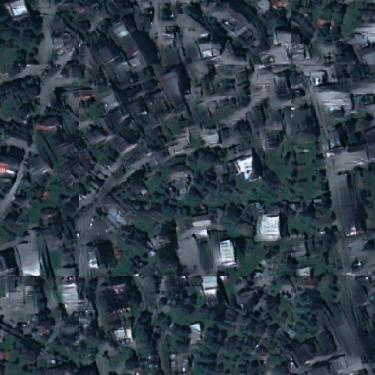}}

\subfloat[]{\includegraphics[width=0.15\linewidth]{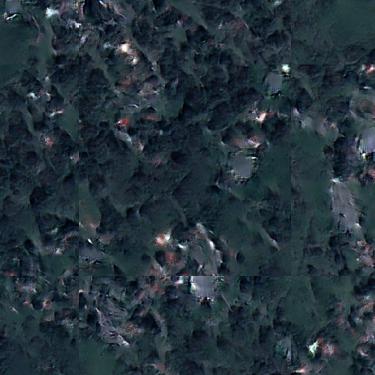}}
\hfill
\subfloat[]{\includegraphics[width=0.15\linewidth]{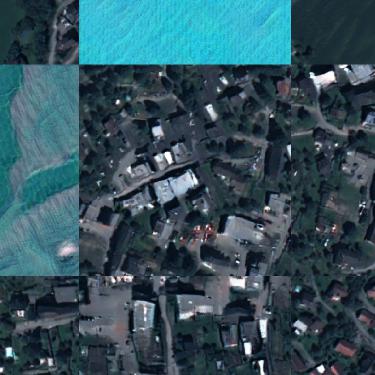}}
\hfill
\subfloat[]{\includegraphics[width=0.15\linewidth]{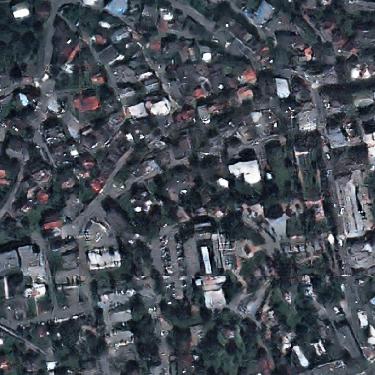}}
\hfill
\subfloat[]{\includegraphics[width=0.15\linewidth]{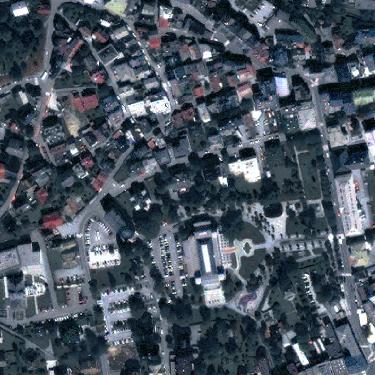}}
\hfill
\subfloat[]{\includegraphics[width=0.15\linewidth]{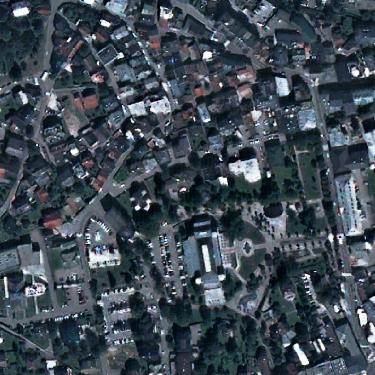}}
\hfill
\subfloat[]{\includegraphics[width=0.15\linewidth]{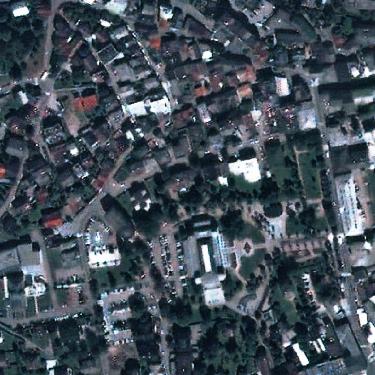}}
\caption{Bad~Ischl and its modified versions to segment Villach. (a)~Bad~Ischl, (b)~gray-world~\cite{buchsbaum1980spatial}, (c)~histogram equalization~\cite{Gonzalez}, (d)~z-score normalization~\cite{aksoy2001feature}, (e)~histogram matching~\cite{Gonzalez}, (f)~UNIT~\cite{liu2017unsupervised}, (g)~MUNIT~\cite{huang2018multimodal}, (h)~DRIT~\cite{lee2018diverse}, (i)~CycleGAN~\cite{zhu2017unpaired}, (j)~ColorMapGAN~\cite{tasar2020colormapgan}, (k)~SemI2I~\cite{tasar2020semi2i}, (l)~DAugNet (ours).}
\label{fig:exp1_bad_ischl}
\end{figure*}

\begin{figure*}
\centering
\subfloat[]{\includegraphics[width=0.15\linewidth]{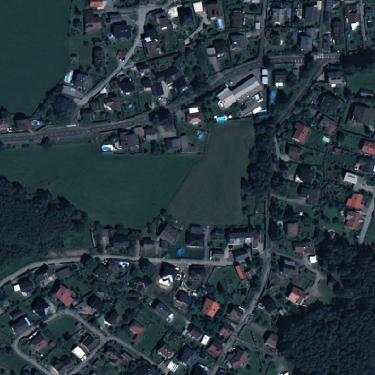}\label{fig:exp1_real_villach}}
\hfill
\subfloat[]{\includegraphics[width=0.15\linewidth]{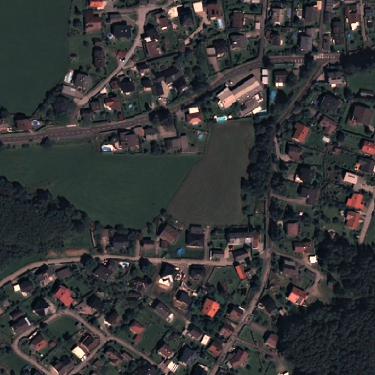}}
\hfill
\subfloat[]{\includegraphics[width=0.15\linewidth]{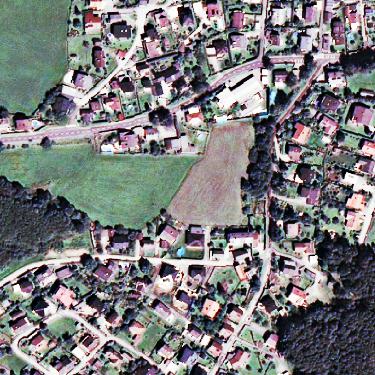}}
\hfill
\subfloat[]{\includegraphics[width=0.15\linewidth]{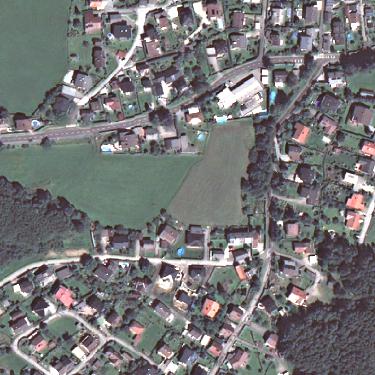}}
\hfill
\subfloat[]{\includegraphics[width=0.15\linewidth]{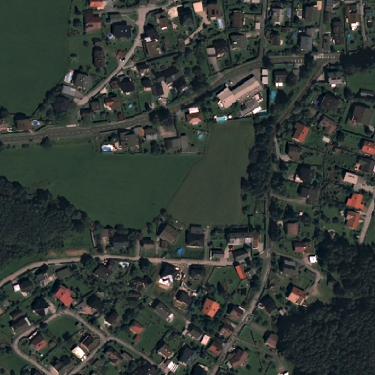}}
\hfill
\subfloat[]{\includegraphics[width=0.15\linewidth]{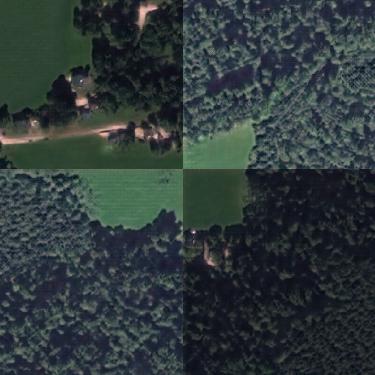}}

\subfloat[]{\includegraphics[width=0.15\linewidth]{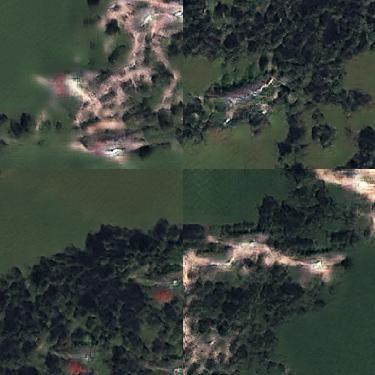}}
\hfill
\subfloat[]{\includegraphics[width=0.15\linewidth]{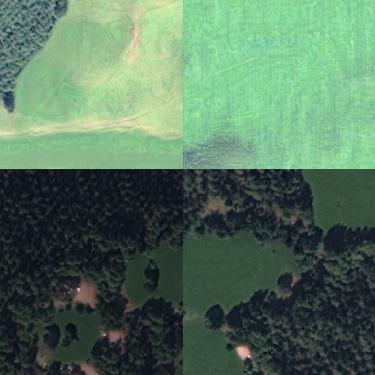}}
\hfill
\subfloat[]{\includegraphics[width=0.15\linewidth]{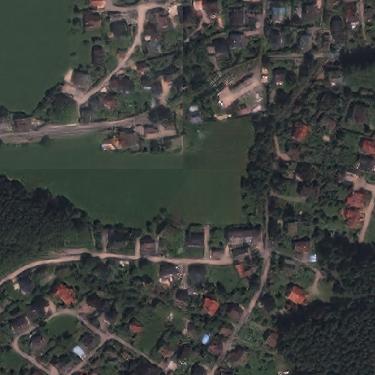}}
\hfill
\subfloat[]{\includegraphics[width=0.15\linewidth]{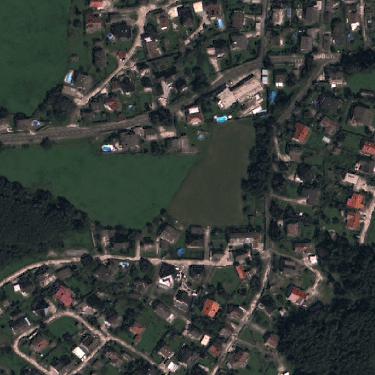}}
\hfill
\subfloat[]{\includegraphics[width=0.15\linewidth]{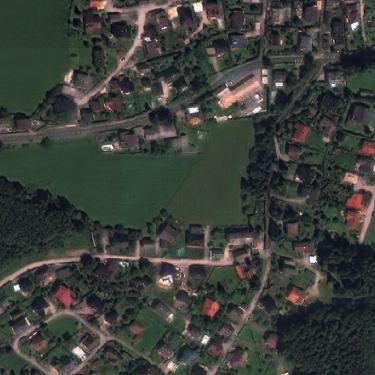}}
\hfill
\subfloat[]{\includegraphics[width=0.15\linewidth]{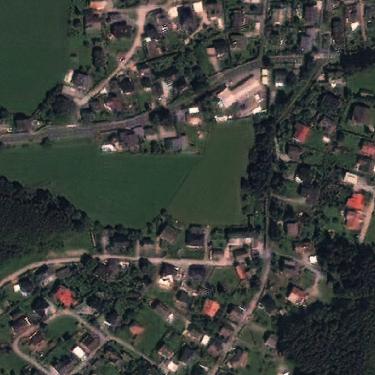}}
\caption{Villach and its modified versions to segment Bad~Ischl. (a)~Villach, (b)~gray-world~\cite{buchsbaum1980spatial}, (c)~histogram equalization~\cite{Gonzalez}, (d)~z-score normalization~\cite{aksoy2001feature}, (e)~histogram matching~\cite{Gonzalez}, (f)~UNIT~\cite{liu2017unsupervised}, (g)~MUNIT~\cite{huang2018multimodal}, (h)~DRIT~\cite{lee2018diverse}, (i)~CycleGAN~\cite{zhu2017unpaired}, (j)~ColorMapGAN~\cite{tasar2020colormapgan}, (k)~SemI2I~\cite{tasar2020semi2i}, (l)~DAugNet (ours).}
\label{fig:exp1_villach}
\end{figure*}

In our first experimental setup, we perform city-to-city adaptation between Villach and Bad~Ischl. A close-up from each city is depicted in Figs.~\ref{fig:exp1_real_bad_ischl} and \ref{fig:exp1_real_villach}. We first use Villach as the source city and Bad~Ischl as the target. We then switch the source and the target cities. In both cases, we assume that the source image is annotated and the target image is unlabeled. We compare our method with thirteen different approaches. The compared methods can be categorized into four groups: naive, adapting the classifier, data standardization, and image-to-image translation. 

The approach belonging to the first category corresponds to training a U-net~\cite{ronneberger2015u} on the source image and segmenting the target city without doing any domain adaptation. As confirmed by Table~\ref{table:iou_exp1}, when there exist a large data distribution difference between the source and the target cities, U-net fails to generate high quality maps. Especially for Bad~Ischl, the quantitative results are extremely poor.

We also apply AdaptSegNet~\cite{tsai2018learning}, which is an approach belonging to the second group, to our data set. This network architecture simultaneously learns from the source city and aligns the features extracted from both cities to train a domain agnostic classifier. The authors use DeepLab~v2~\cite{chen2018deeplab} as the classifier. The feature alignment can be executed in one or multiple layers. In Table~\ref{table:iou_exp1}, AdaptSegNet Single and AdaptSegNet Multi respectively correspond to the proposed method when the alignment is done in the last and in the last two layers before upsampling. However, the quantitative results prove that this method does not exhibit a good performance on domain adaptation of satellite images.

Among the methods based on data standardization, we compare our approach with gray-world algorithm~\cite{buchsbaum1980spatial}, histogram equalization~\cite{Gonzalez}, and z-score normalization~\cite{aksoy2001feature}. We first standardize images from both cities. We then train a U-net on the standardized source image and segment the standardized target data. Although these standardization methods enable U-net to better generalize, the improvement is unsatisfactory. As shown in Figs.~\ref{fig:exp1_bad_ischl} and \ref{fig:exp1_villach}, while the data distributions of the images get close to each other with these approaches, we still observe some differences. 

To evaluate image-to-image translation based methods, we first a generate target stylized fake source city. Afterwards, we train a U-net on the fake source city and evaluate it on the target city. We provide the results for histogram matching~\cite{Gonzalez}, UNIT~\cite{liu2017unsupervised}, MUNIT~\cite{huang2018multimodal}, DRIT~\cite{lee2018diverse}, CycleGAN~\cite{zhu2017unpaired}, ColorMapGAN~\cite{tasar2020colormapgan}, and SemI2I~\cite{tasar2020semi2i}. As illustrated in Figs.~\ref{fig:exp1_bad_ischl} and \ref{fig:exp1_villach}, the fake images generated by UNIT~\cite{liu2017unsupervised}, MUNIT~\cite{huang2018multimodal}, and DRIT~\cite{lee2018diverse} are semantically inconsistent with the real source images. Therefore, U-net learns from data, where image and the ground-truth do not match. As a consequence, the quantitative results for these approaches are poor. Histogram matching does not take into account the contextual information. For example, fake image generated by histogram matching from Bad~Ischl contains a lot of buildings with violet or cyan rooftops. However, such buildings do not exist in Villach. Since the fake source image is not representative for the target image, the performance of this approach is unsatisfactory. CycleGAN generates blurry fake source images, which negatively affects the performance of U-net. ColorMapGAN and SemI2I are better performers than the other ones.

When we train the multi-domain style transfer part of our solution, we set $\lambda_1$, $\lambda_2$, $\lambda_3$, and $\lambda_4$ parameters in Eqs.~\ref{eq:g_loss} and \ref{eq:d_loss} to 1, 10, 10, and 100, respectively. We have found these values empirically. We train our style transfer method for 25 epochs. The learning rate for the initial 15 epochs is 0.0001. We reduce it gradually in the rest of the epochs as:
\begin{equation}\label{eq:lr}
\text{LR} = \text{0.0001} \times \frac{\text{num\_epochs} - \text{epoch\_no}}{\text{num\_epochs} - \text{decay\_epoch}}, 
\end{equation}
where LR, num\_epochs, epoch\_no, and decay\_epoch denote the current learning rate, total number of epochs, current epoch no, and the epoch no where we start reducing the learning rate. As the optimizer, we use Adam algorithm~\cite{kingma2014adam}. We set the beta coefficients of Adam to 0.5 and 0.999, respectively.  When we train DAugNet, our data augmentor generates only target stylized image batches, since there are only two cities in this experimental setup. Note that any network architecture can be used as the classifier of DAugNet. We prefer to use U-net. We train DAugNet for 35 epochs with 0.0001 learning rate. Except for AdapSegNet, we train U-net for all the compared methods for 35 epochs with the same learning rate. When training DAugNet and U-net for the compared methods, we set $\lambda_5$ and $\lambda_6$ in Eq.~\ref{eq:class_loss} to 0.25 and 0.75, respectively. We sample a batch of 32 training patches and perform online data augmentation with random flips and rotations. Table~\ref{table:iou_exp1} proves that DAugNet performs the best in most cases.

\begin{figure*}
\centering

\begin{tabular}{p{0.1em}p{0.1em}c@{\hskip3pt}c@{\hskip3pt}c@{\hskip3pt}c@{\hskip3pt}c@{\hskip3pt}c@{\hskip3pt}}
\multicolumn{8}{c}{\large{\textbf{~~~~~~~~~STYLE}}} \\
& & Villach & Sankt P{\"o}lten & Bourges & Lille & Bad~Ischl & Vaduz \\
\multirow{35}{*}{\rotatebox[origin=c]{90}{\large{\textbf{CONTENT}}}}&
\rotatebox[origin=c]{90}{Villach}&
\raisebox{-.5\height}{\frame{\includegraphics[width=0.15\linewidth]{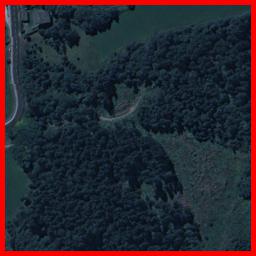}}} &
\raisebox{-.5\height}{\frame{\includegraphics[width=0.15\linewidth]{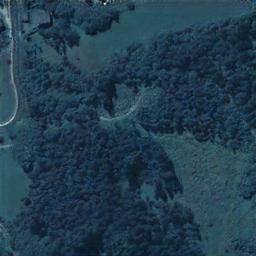}}} &
\raisebox{-.5\height}{\frame{\includegraphics[width=0.15\linewidth]{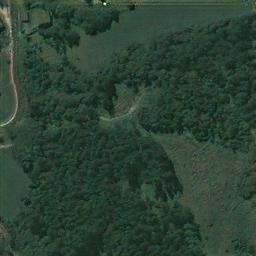}}} &
\raisebox{-.5\height}{\frame{\includegraphics[width=0.15\linewidth]{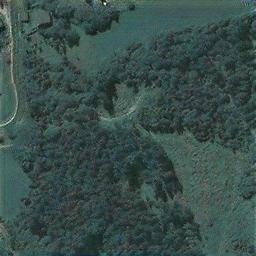}}} &
\raisebox{-.5\height}{\frame{\includegraphics[width=0.15\linewidth]{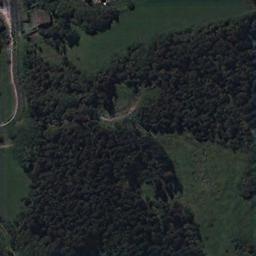}}} &
\raisebox{-.5\height}{\frame{\includegraphics[width=0.15\linewidth]{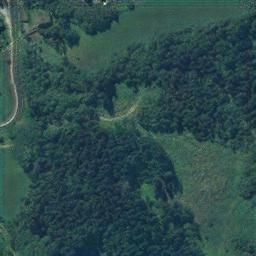}}}\\
\\[-3.5mm]
&\rotatebox[origin=c]{90}{Sankt P{\"o}lten}&
\raisebox{-.5\height}{\frame{\includegraphics[width=0.15\linewidth]{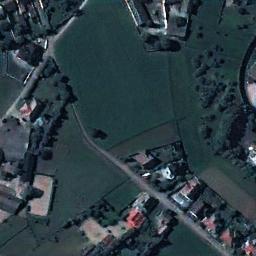}}} &
\raisebox{-.5\height}{\frame{\includegraphics[width=0.15\linewidth]{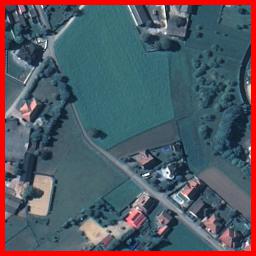}}} &
\raisebox{-.5\height}{\frame{\includegraphics[width=0.15\linewidth]{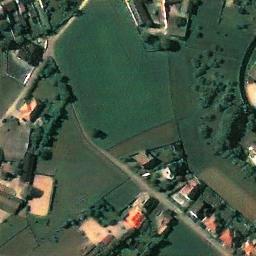}}} &
\raisebox{-.5\height}{\frame{\includegraphics[width=0.15\linewidth]{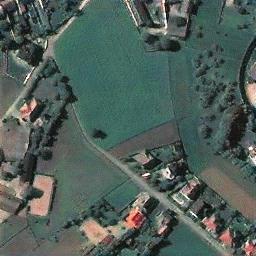}}} &
\raisebox{-.5\height}{\frame{\includegraphics[width=0.15\linewidth]{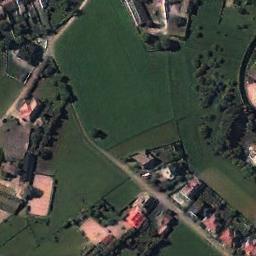}}} &
\raisebox{-.5\height}{\frame{\includegraphics[width=0.15\linewidth]{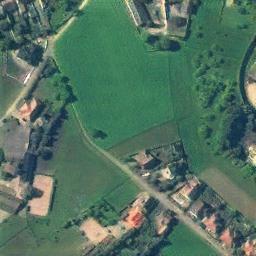}}}\\
\\[-3.5mm]
&\rotatebox[origin=c]{90}{Bourges}&
\raisebox{-.5\height}{\frame{\includegraphics[width=0.15\linewidth]{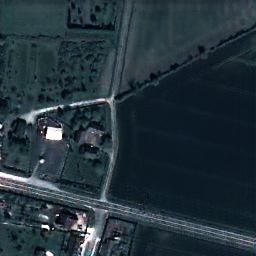}}} &
\raisebox{-.5\height}{\frame{\includegraphics[width=0.15\linewidth]{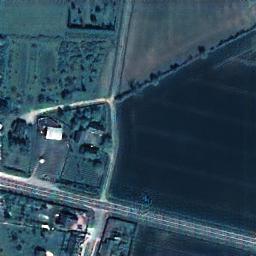}}} &
\raisebox{-.5\height}{\frame{\includegraphics[width=0.15\linewidth]{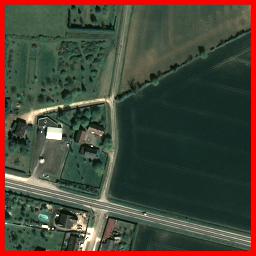}}} &
\raisebox{-.5\height}{\frame{\includegraphics[width=0.15\linewidth]{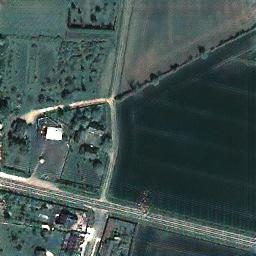}}} &
\raisebox{-.5\height}{\frame{\includegraphics[width=0.15\linewidth]{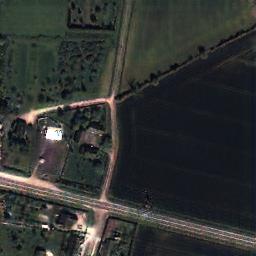}}} &
\raisebox{-.5\height}{\frame{\includegraphics[width=0.15\linewidth]{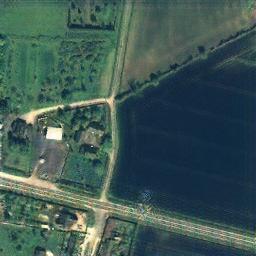}}}\\
\\[-3.5mm]
&\rotatebox[origin=c]{90}{Lille}&
\raisebox{-.5\height}{\frame{\includegraphics[width=0.15\linewidth]{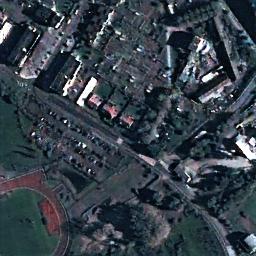}}} &
\raisebox{-.5\height}{\frame{\includegraphics[width=0.15\linewidth]{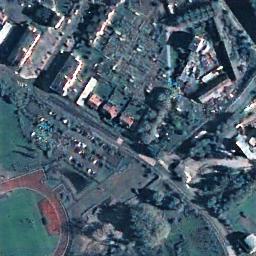}}} &
\raisebox{-.5\height}{\frame{\includegraphics[width=0.15\linewidth]{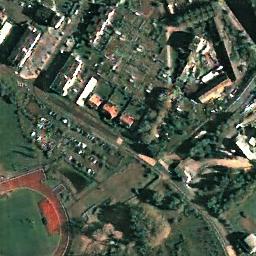}}} &
\raisebox{-.5\height}{\frame{\includegraphics[width=0.15\linewidth]{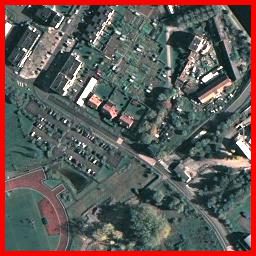}}} &
\raisebox{-.5\height}{\frame{\includegraphics[width=0.15\linewidth]{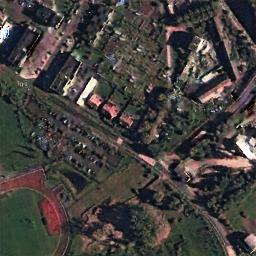}}} &
\raisebox{-.5\height}{\frame{\includegraphics[width=0.15\linewidth]{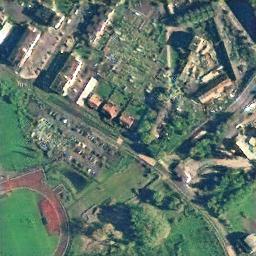}}}\\
\\[-3.5mm]
&\rotatebox[origin=c]{90}{Bad~Ischl}&
\raisebox{-.5\height}{\frame{\includegraphics[width=0.15\linewidth]{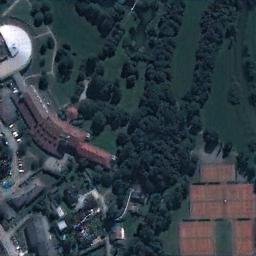}}} &
\raisebox{-.5\height}{\frame{\includegraphics[width=0.15\linewidth]{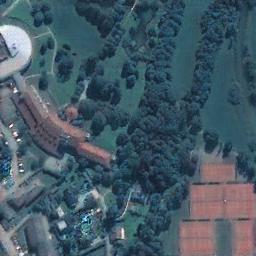}}} &
\raisebox{-.5\height}{\frame{\includegraphics[width=0.15\linewidth]{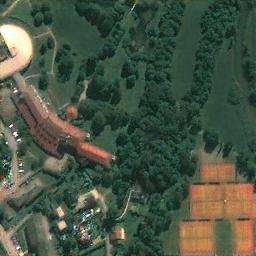}}} &
\raisebox{-.5\height}{\frame{\includegraphics[width=0.15\linewidth]{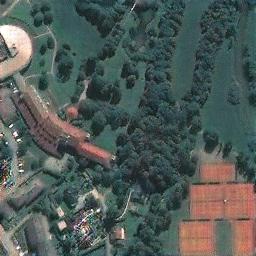}}} &
\raisebox{-.5\height}{\frame{\includegraphics[width=0.15\linewidth]{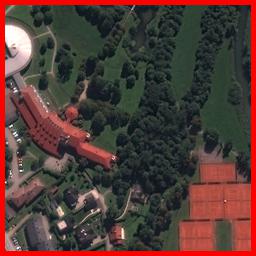}}} &
\raisebox{-.5\height}{\frame{\includegraphics[width=0.15\linewidth]{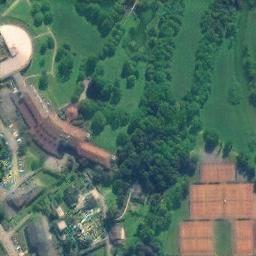}}}\\
\\[-3.5mm]
&\rotatebox[origin=c]{90}{Vaduz}&
\raisebox{-.5\height}{\frame{\includegraphics[width=0.15\linewidth]{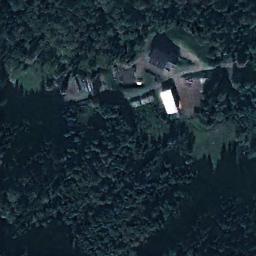}}} &
\raisebox{-.5\height}{\frame{\includegraphics[width=0.15\linewidth]{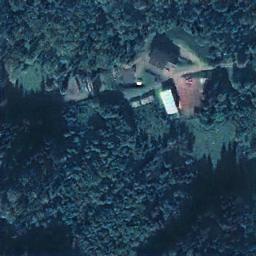}}} &
\raisebox{-.5\height}{\frame{\includegraphics[width=0.15\linewidth]{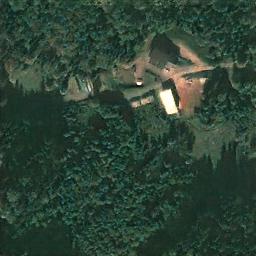}}} &
\raisebox{-.5\height}{\frame{\includegraphics[width=0.15\linewidth]{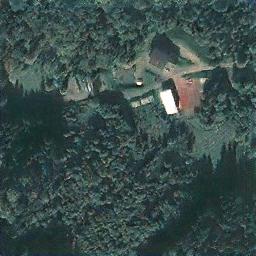}}} &
\raisebox{-.5\height}{\frame{\includegraphics[width=0.15\linewidth]{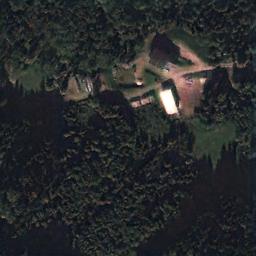}}} &
\raisebox{-.5\height}{\frame{\includegraphics[width=0.15\linewidth]{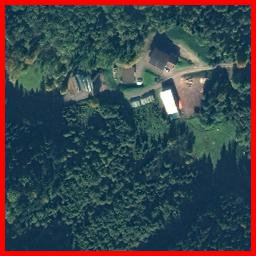}}}\\
\end{tabular}
\caption{Style transfer between the images used in the second experiment. The cells with red bounding boxes are real data. The rest of the cells represent the fake data generated by our multi-domain style transfer approach.}
\label{fig:exp2_md_style_transfer}
\end{figure*}

\begin{table*}
\centering
\caption{IoU scores for the second experiment (multi-source and multi-target adaptation).}
\label{table:iou_exp2}
\begin{tabular}{||c|c||cccc||cccc||}
\hline
\multicolumn{2}{||c||}{\multirow{2}{*}{\textbf{Method}}} & \multicolumn{4}{c||}{\textbf{Bad Ischl}} & \multicolumn{4}{c||}{\textbf{Vaduz}} \\
\cline{3-10}
\multicolumn{1}{||c}{} &  & \textbf{Building} & \textbf{Road} & \textbf{Tree} & \textbf{Overall} & \textbf{Building} & \textbf{Road} & \textbf{Tree} & \textbf{Overall} \\
\hline

\multicolumn{2}{||c||}{U-net~\cite{ronneberger2015u}}  & 35.48 & 26.36 & 51.24 & 37.69 & 12.49 & 18.68 & 58.34 & 29.84 \\  
\hline
U-net trained & Gray-world~\cite{buchsbaum1980spatial}      & 50.30 & 43.00 & 70.41 & 54.57 & 19.45 & 19.59 & 18.85 & 19.29 \\
on data       & Histogram Equalization~\cite{Gonzalez}  & 42.32 & 38.21 & 75.16 & 51.90 & 19.32 & 20.64 & 68.50 & 36.15 \\
generated by  & Z-score normalization~\cite{aksoy2001feature}   & 39.77 & 41.86 & 78.53 & 53.39 & 15.40 & 24.16 & 58.94 & 32.83 \\
\hline
\multicolumn{2}{||c||}{DAugNet} & \textbf{59.70} & \textbf{49.05} & \textbf{83.39} & \textbf{64.05} & \textbf{51.83} & \textbf{34.41} & \textbf{69.74} & \textbf{51.99} \\
\cline{3-10}
\hline
\end{tabular}
\end{table*}

\begin{figure*}
\centering
\begin{tabular}{p{0.1em}c@{\hspace{0.15em}}c@{\hspace{0.15em}}c@{\hspace{0.15em}}c@{\hspace{0.15em}}c@{\hspace{0.15em}}c@{\hspace{0.15em}}c@{\hspace{0.15em}}}

& Image & Ground-truth & U-net~\cite{ronneberger2015u} & Gray-world~\cite{buchsbaum1980spatial} & Hist.~eq.~\cite{Gonzalez} & Z-score nor.~\cite{aksoy2001feature} & DAugNet (ours) \\
\\[-3.5mm]
\rotatebox[origin=c]{90}{Bad~Ischl}&
\raisebox{-.5\height}{\frame{\includegraphics[width=0.135\linewidth]{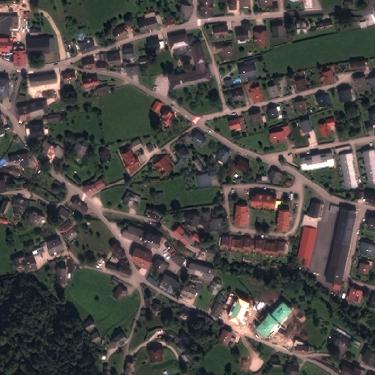}}} &
\raisebox{-.5\height}{\frame{\includegraphics[width=0.135\linewidth]{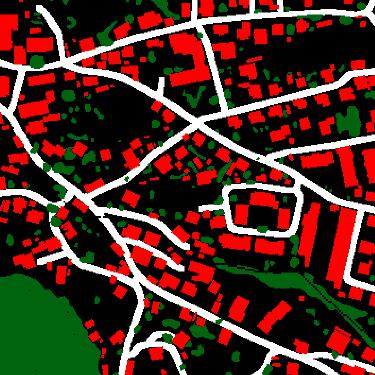}}} &
\raisebox{-.5\height}{\frame{\includegraphics[width=0.135\linewidth]{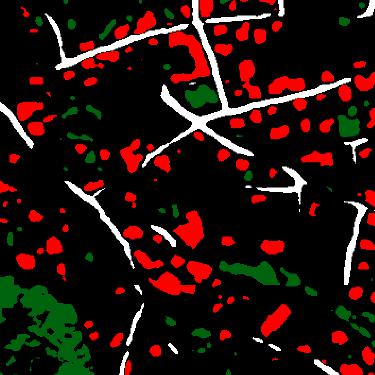}}} &
\raisebox{-.5\height}{\frame{\includegraphics[width=0.135\linewidth]{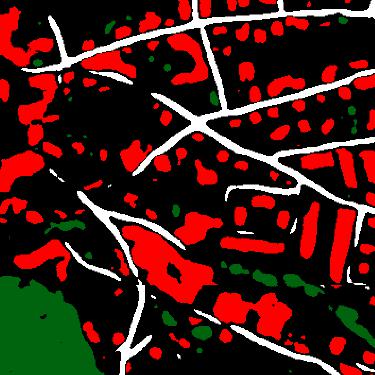}}} &
\raisebox{-.5\height}{\frame{\includegraphics[width=0.135\linewidth]{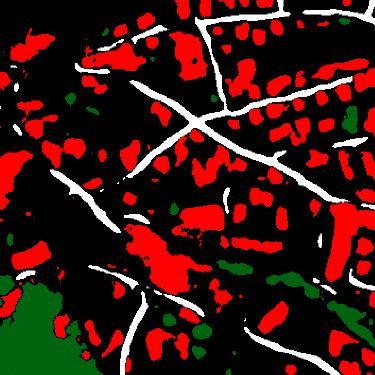}}} &
\raisebox{-.5\height}{\frame{\includegraphics[width=0.135\linewidth]{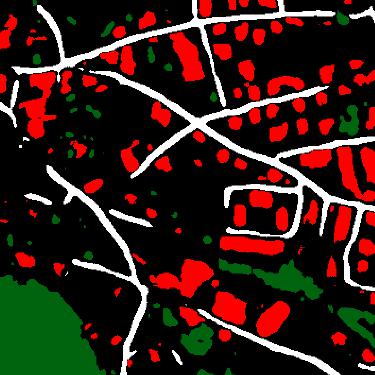}}} & 
\raisebox{-.5\height}{\frame{\includegraphics[width=0.135\linewidth]{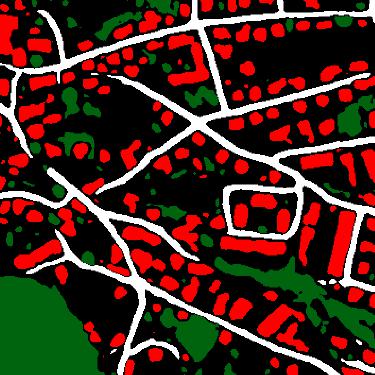}}}\\
\\[-3.5mm]

\rotatebox[origin=c]{90}{Vaduz}&
\raisebox{-.5\height}{\frame{\includegraphics[width=0.135\linewidth]{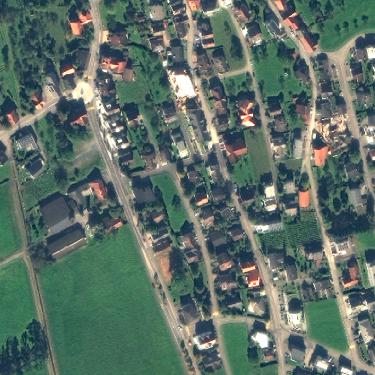}}} &
\raisebox{-.5\height}{\frame{\includegraphics[width=0.135\linewidth]{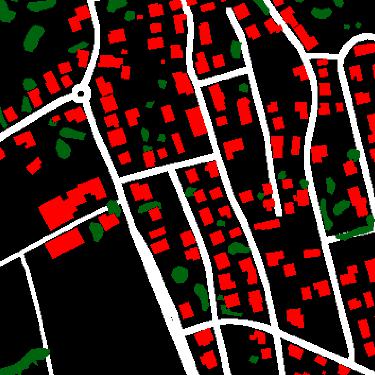}}} &
\raisebox{-.5\height}{\frame{\includegraphics[width=0.135\linewidth]{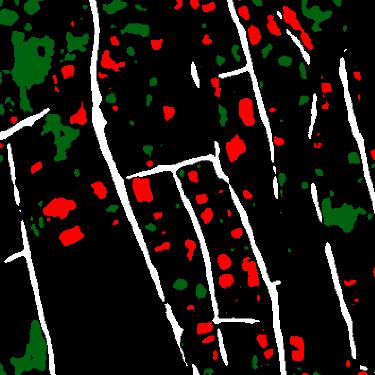}}} &
\raisebox{-.5\height}{\frame{\includegraphics[width=0.135\linewidth]{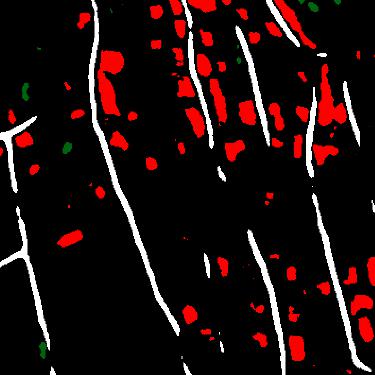}}} &
\raisebox{-.5\height}{\frame{\includegraphics[width=0.135\linewidth]{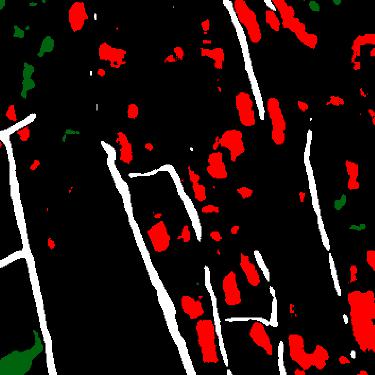}}} &
\raisebox{-.5\height}{\frame{\includegraphics[width=0.135\linewidth]{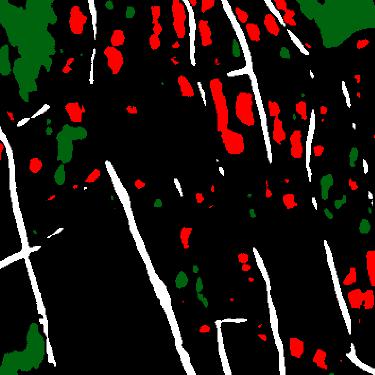}}} & 
\raisebox{-.5\height}{\frame{\includegraphics[width=0.135\linewidth]{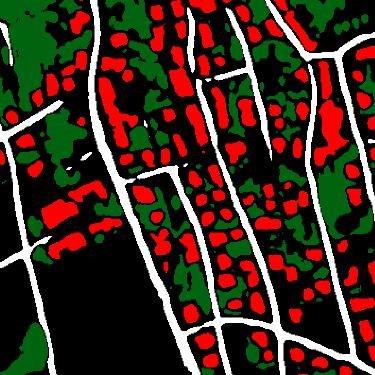}}}\\
\end{tabular}
\caption{Target images used in the second experiment, their ground-truth, and the predictions. Red, white, and green pixels represent building, road, and tree classes, respectively. The pixels in black do not belong to any class.}
\label{fig:exp2_preds}
\end{figure*}

\subsection{Multi-source and Multi-target Domain Adaptation}
To evaluate how well our approach performs on Multi-source and Multi-target adaptation setting, we choose Villach, Sankt~P{\"o}lten, Bourges, and Lille as the source, and Bad~Ischl and Vaduz as the target cities. In this setup, we train our initial multi-domain style transfer approach for 200 epochs. We use Adam optimizer with the initial learning of 0.0001 and the same beta coefficients used in the previous experiment. Starting from the 100$^{th}$ epoch, we reduce the learning rate via Eq.~\ref{eq:lr}. We train DAugNet as described in Sec.~\ref{sec:daugnet} for 100 epochs. We compare our framework with the same standardization approaches explained in the previous experimental setup. Once all the data are standardized, we train a U-net on the standardized source images for 100 epochs and report the results on the standardized target cities. 

Table~\ref{table:iou_exp2} reports the IoU scores for multi-source and multi-target experimental setup. Our first observation is that adding more source cities with different data distributions bridges the domain gap between the source and the target cities. Thus, the classifier better generalizes to new geographic locations. For instance, although we use Bad~Ischl as the target city both in this experiment and in the previous one, the quantitative results of each method for this city are significantly higher in this setting. Secondly, in some cases, the performances of the standardization algorithms are unstable. All of these algorithms allow the classifier to exhibit a considerably better performance on Bad~Ischl, whereas the quantitative results for Vaduz are either roughly on par with or even worse than naive U-net. As depicted in Fig.~\ref{fig:exp2_md_style_transfer}, our multi-domain style transfer method is able to stylize each city as another one. As mentioned earlier, diversifying a batch of image patches when training DAugNet with the help of the data augmentor enables DAugNet to learn from data that are not only representative for the source cities but also representative for the target cities. Hence, DAugNet achieves a better performance than the others. Fig~\ref{fig:exp2_preds} shows a close-up from each city, its ground-truth, and the predictions.

\begin{figure*}
\centering
\begin{tabular}{p{0.1em}p{0.1em}c@{\hspace{0.15em}}c@{\hspace{0.15em}}c@{\hspace{0.15em}}c@{\hspace{0.15em}}c@{\hspace{0.15em}}c@{\hspace{0.15em}}c@{\hspace{0.15em}}c@{\hspace{0.15em}}c@{\hspace{0.15em}}c@{\hspace{0.15em}}}

\multicolumn{12}{c}{\large{\textbf{~~~~~~~~~STYLE}}} \\

& & B{\'e}ziers & Salzburg & Albi & Leibnitz & Villach & S. P{\"o}lten & Bourges & Lille & Bad Ischl & Vaduz \\

\multirow{14}{*}{\rotatebox[origin=c]{90}{\large{\textbf{CONTENT}}}}
& \rotatebox[origin=c]{90}{B{\'e}ziers} &
\raisebox{-.5\height}{\frame{\includegraphics[width=0.09\linewidth]{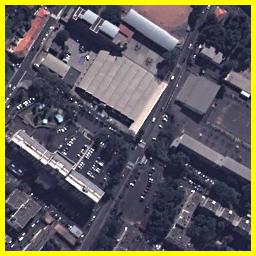}}} &
\raisebox{-.5\height}{\frame{\includegraphics[width=0.09\linewidth]{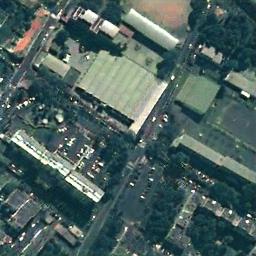}}} &
\raisebox{-.5\height}{\frame{\includegraphics[width=0.09\linewidth]{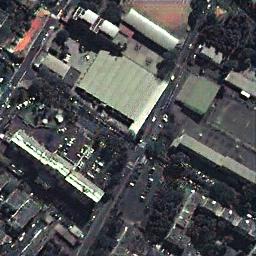}}} &
\raisebox{-.5\height}{\frame{\includegraphics[width=0.09\linewidth]{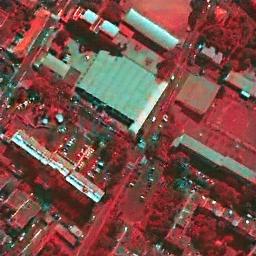}}} &
\raisebox{-.5\height}{\frame{\includegraphics[width=0.09\linewidth]{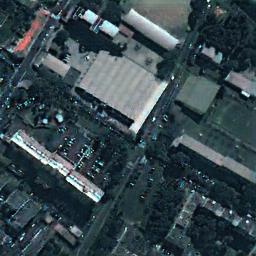}}} &
\raisebox{-.5\height}{\frame{\includegraphics[width=0.09\linewidth]{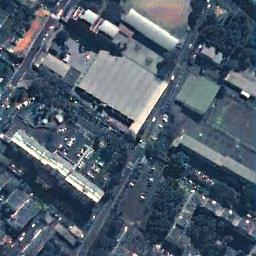}}} & 
\raisebox{-.5\height}{\frame{\includegraphics[width=0.09\linewidth]{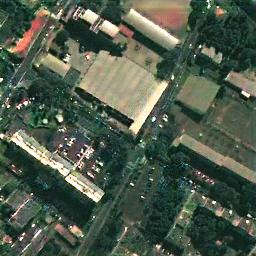}}} &
\raisebox{-.5\height}{\frame{\includegraphics[width=0.09\linewidth]{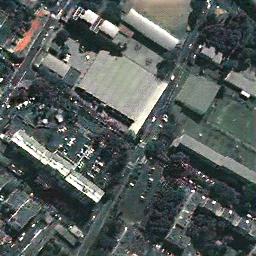}}} &
\raisebox{-.5\height}{\frame{\includegraphics[width=0.09\linewidth]{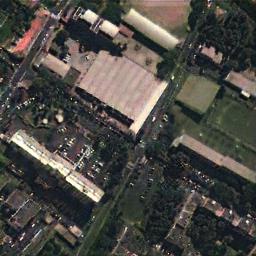}}} &
\raisebox{-.5\height}{\frame{\includegraphics[width=0.09\linewidth]{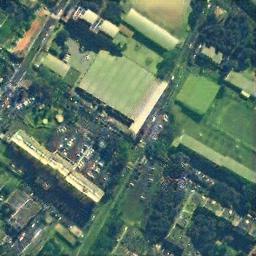}}}\\
\\[-3.5mm]

& \rotatebox[origin=c]{90}{Salzburg} &
\raisebox{-.5\height}{\frame{\includegraphics[width=0.09\linewidth]{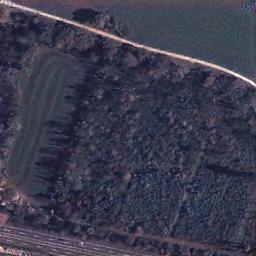}}} &
\raisebox{-.5\height}{\frame{\includegraphics[width=0.09\linewidth]{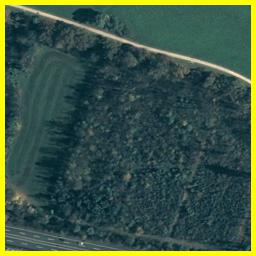}}} &
\raisebox{-.5\height}{\frame{\includegraphics[width=0.09\linewidth]{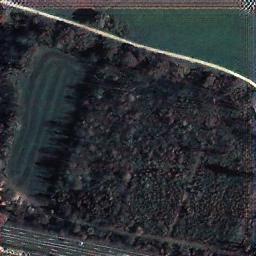}}} &
\raisebox{-.5\height}{\frame{\includegraphics[width=0.09\linewidth]{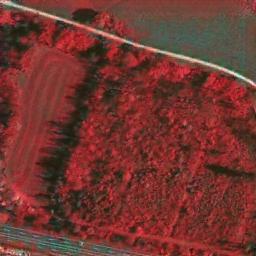}}} &
\raisebox{-.5\height}{\frame{\includegraphics[width=0.09\linewidth]{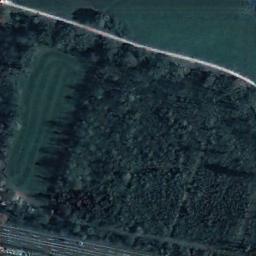}}} &
\raisebox{-.5\height}{\frame{\includegraphics[width=0.09\linewidth]{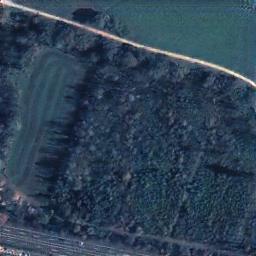}}} & 
\raisebox{-.5\height}{\frame{\includegraphics[width=0.09\linewidth]{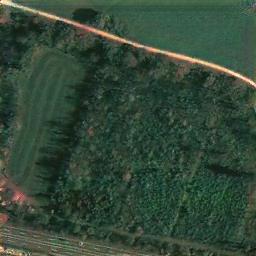}}} &
\raisebox{-.5\height}{\frame{\includegraphics[width=0.09\linewidth]{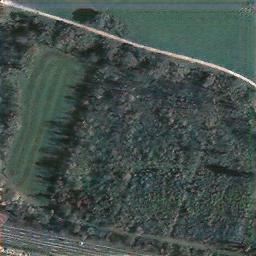}}} &
\raisebox{-.5\height}{\frame{\includegraphics[width=0.09\linewidth]{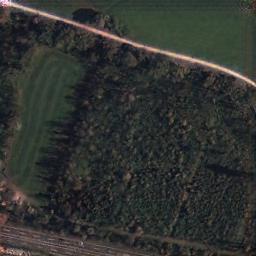}}} &
\raisebox{-.5\height}{\frame{\includegraphics[width=0.09\linewidth]{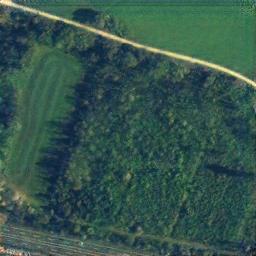}}}\\
\\[-3.5mm]

& \rotatebox[origin=c]{90}{Albi} & 
\raisebox{-.5\height}{\frame{\includegraphics[width=0.09\linewidth]{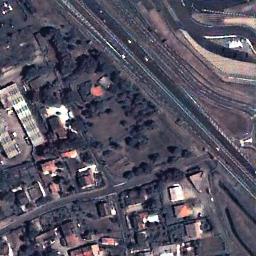}}} &
\raisebox{-.5\height}{\frame{\includegraphics[width=0.09\linewidth]{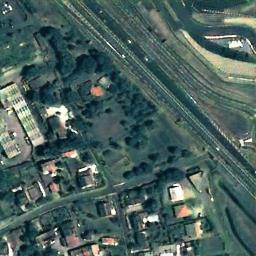}}} &
\raisebox{-.5\height}{\frame{\includegraphics[width=0.09\linewidth]{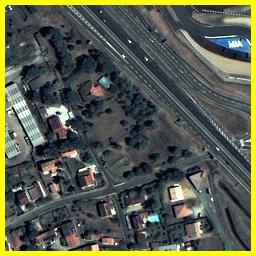}}} &
\raisebox{-.5\height}{\frame{\includegraphics[width=0.09\linewidth]{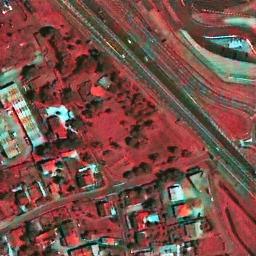}}} &
\raisebox{-.5\height}{\frame{\includegraphics[width=0.09\linewidth]{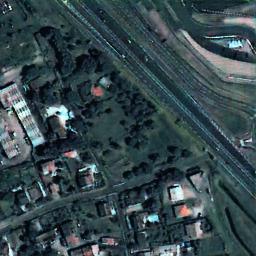}}} &
\raisebox{-.5\height}{\frame{\includegraphics[width=0.09\linewidth]{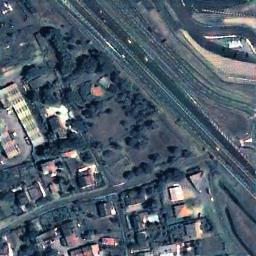}}} & 
\raisebox{-.5\height}{\frame{\includegraphics[width=0.09\linewidth]{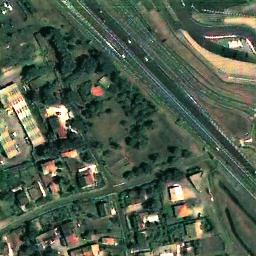}}} &
\raisebox{-.5\height}{\frame{\includegraphics[width=0.09\linewidth]{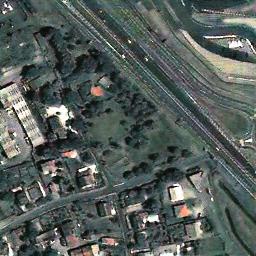}}} &
\raisebox{-.5\height}{\frame{\includegraphics[width=0.09\linewidth]{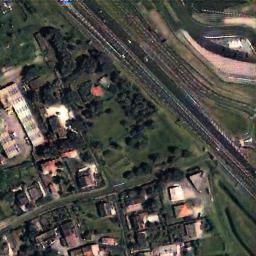}}} &
\raisebox{-.5\height}{\frame{\includegraphics[width=0.09\linewidth]{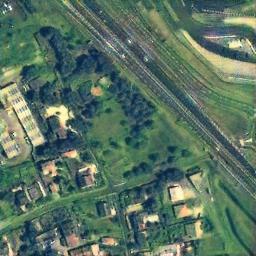}}}\\
\\[-3.5mm]

& \rotatebox[origin=c]{90}{Leibnitz} &
\raisebox{-.5\height}{\frame{\includegraphics[width=0.09\linewidth]{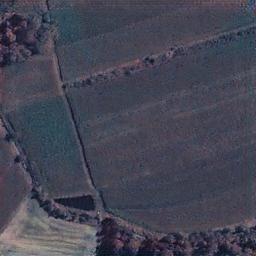}}} &
\raisebox{-.5\height}{\frame{\includegraphics[width=0.09\linewidth]{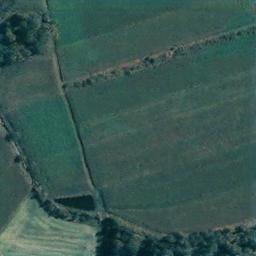}}} &
\raisebox{-.5\height}{\frame{\includegraphics[width=0.09\linewidth]{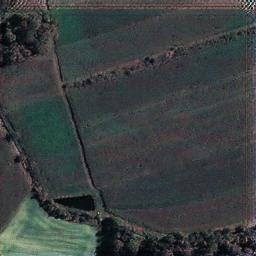}}} &
\raisebox{-.5\height}{\frame{\includegraphics[width=0.09\linewidth]{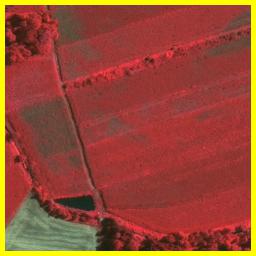}}} &
\raisebox{-.5\height}{\frame{\includegraphics[width=0.09\linewidth]{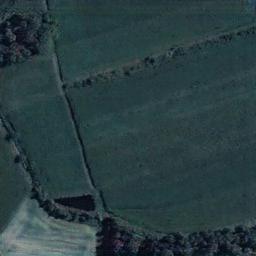}}} &
\raisebox{-.5\height}{\frame{\includegraphics[width=0.09\linewidth]{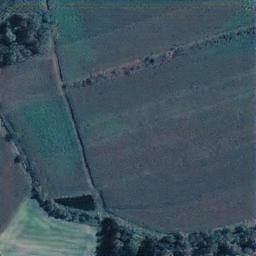}}} & 
\raisebox{-.5\height}{\frame{\includegraphics[width=0.09\linewidth]{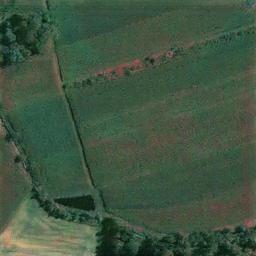}}} &
\raisebox{-.5\height}{\frame{\includegraphics[width=0.09\linewidth]{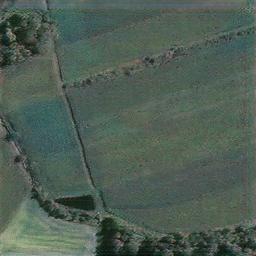}}} &
\raisebox{-.5\height}{\frame{\includegraphics[width=0.09\linewidth]{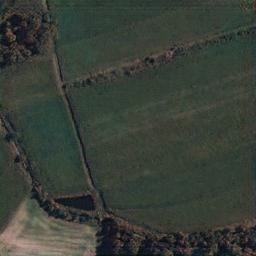}}} &
\raisebox{-.5\height}{\frame{\includegraphics[width=0.09\linewidth]{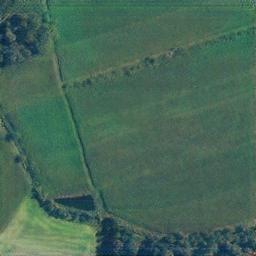}}}\\
\end{tabular}
\caption{Style transfer between newly added cities in the third experiment and all the cities. The cells with yellow bounding boxes represent the additional real images. The rest of the cells show the fake images generated by our multi-domain style transfer approach.}
\label{fig:exp3_md_style_transfer}
\end{figure*}

\begin{table*}
\centering
\caption{IoU scores for the third experiment (multi-source, multi-target, and life-long adaptation).}
\label{table:iou_exp3}
\begin{tabular}{||c|c||cccc||cccc||cccc||}
\hline
\multicolumn{2}{||c||}{\multirow{2}{*}{\textbf{Method}}} & \multicolumn{4}{c||}{\textbf{Bad Ischl}} & \multicolumn{4}{c||}{\textbf{Vaduz}} & \multicolumn{4}{c||}{\textbf{Leibnitz}} \\
\cline{3-14}
\multicolumn{1}{||c}{} &  & \textbf{Build.} & \textbf{Road} & \textbf{Tree} & \textbf{Overall} & \textbf{Build.} & \textbf{Road} & \textbf{Tree} & \textbf{Overall} & \textbf{Build.} & \textbf{Road} & \textbf{Tree} & \textbf{Overall} \\
\hline

\multicolumn{2}{||c||}{U-net~\cite{ronneberger2015u}}  &  51.86 & 39.33 & 79.80 & 56.99 & 38.04 & 34.33 & \textbf{75.88} & 49.42 & 2.29 & 1.13 & 0.04 & 1.15  \\  
\hline
\multirow{3}{*}{\rotatebox{90}{Unet on}}
& Gray-world~\cite{buchsbaum1980spatial} & 59.59 & 47.95 & 68.72 & 58.75 & 26.04 & 30.98 & 45.37 & 34.13 & 0.01 & 0.10 & 21.59 & 7.23  \\
& Hist.~Eq.~\cite{Gonzalez}              & 44.03 & 39.94 & 78.56 & 54.18 & 19.19 & 23.64 & 71.11 & 37.98 & 1.23 & 0.42 & 55.93 & 19.19 \\
& Z-score norm.~\cite{aksoy2001feature}  & 46.58 & 44.62 & 75.25 & 55.48 & 29.08 & 28.97 & 48.38 & 35.48 & 0.10 & 1.14 & 11.80 & 4.35 \\
\hline
\multicolumn{2}{||c||}{DAugNet} & \textbf{60.52} & \textbf{53.08} & \textbf{82.87} & \textbf{65.49} & \textbf{53.03} & \textbf{39.44} & 71.77 & \textbf{54.75} & \textbf{43.31} & \textbf{34.28} & \textbf{73.18} & \textbf{50.26} \\
\cline{3-10}
\hline
\end{tabular}
\end{table*}

\begin{figure*}
\centering
\begin{tabular}{p{0.1em}c@{\hspace{0.15em}}c@{\hspace{0.15em}}c@{\hspace{0.15em}}c@{\hspace{0.15em}}c@{\hspace{0.15em}}c@{\hspace{0.15em}}c@{\hspace{0.15em}}}

& Image & Ground-truth & U-net~\cite{ronneberger2015u} & Gray-world~\cite{buchsbaum1980spatial} & Hist.~eq.~\cite{Gonzalez} & Z-score nor.~\cite{aksoy2001feature} & DAugNet (ours) \\
\\[-3.5mm]
\rotatebox[origin=c]{90}{Bad~Ischl}&
\raisebox{-.5\height}{\frame{\includegraphics[width=0.135\linewidth]{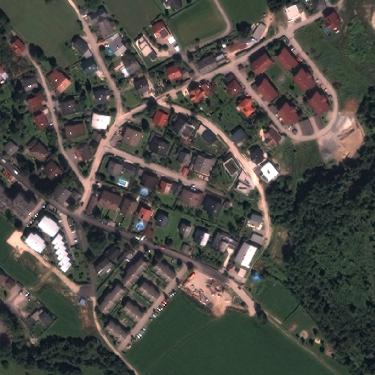}}} &
\raisebox{-.5\height}{\frame{\includegraphics[width=0.135\linewidth]{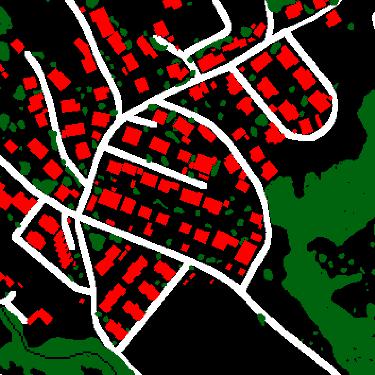}}} &
\raisebox{-.5\height}{\frame{\includegraphics[width=0.135\linewidth]{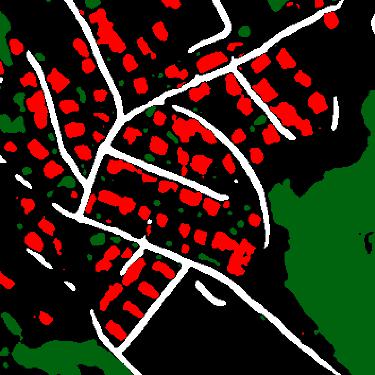}}} &
\raisebox{-.5\height}{\frame{\includegraphics[width=0.135\linewidth]{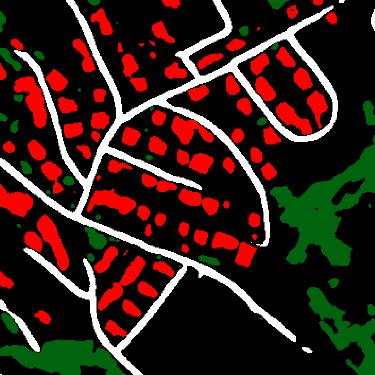}}} &
\raisebox{-.5\height}{\frame{\includegraphics[width=0.135\linewidth]{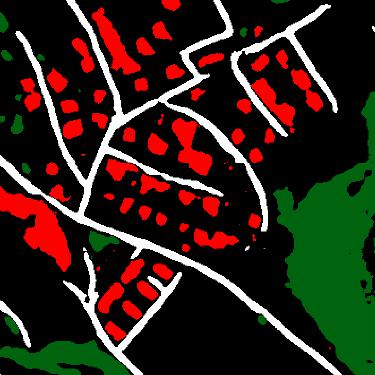}}} &
\raisebox{-.5\height}{\frame{\includegraphics[width=0.135\linewidth]{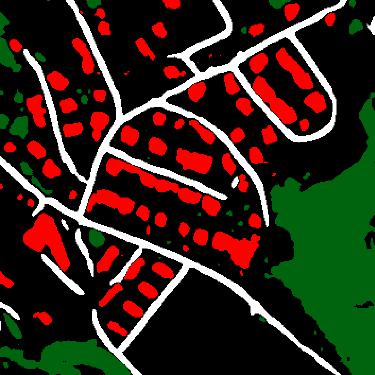}}} & 
\raisebox{-.5\height}{\frame{\includegraphics[width=0.135\linewidth]{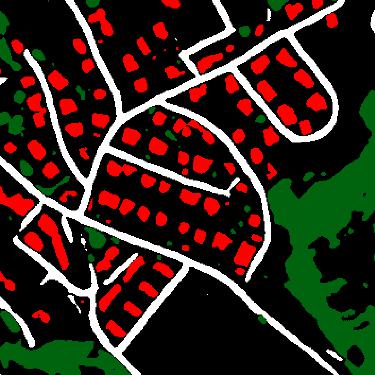}}}\\
\\[-3.5mm]

\rotatebox[origin=c]{90}{Vaduz}&
\raisebox{-.5\height}{\frame{\includegraphics[width=0.135\linewidth]{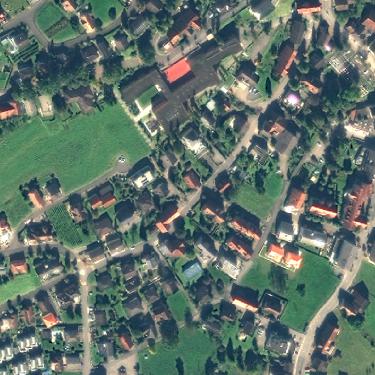}}} &
\raisebox{-.5\height}{\frame{\includegraphics[width=0.135\linewidth]{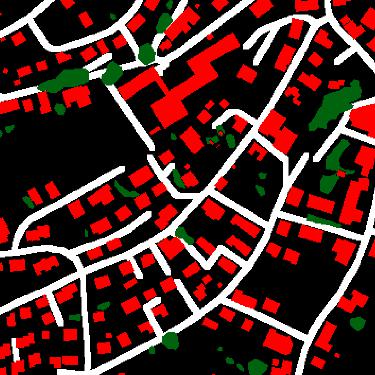}}} &
\raisebox{-.5\height}{\frame{\includegraphics[width=0.135\linewidth]{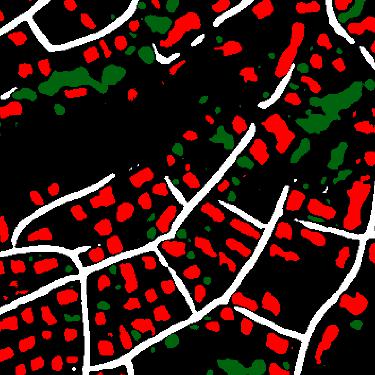}}} &
\raisebox{-.5\height}{\frame{\includegraphics[width=0.135\linewidth]{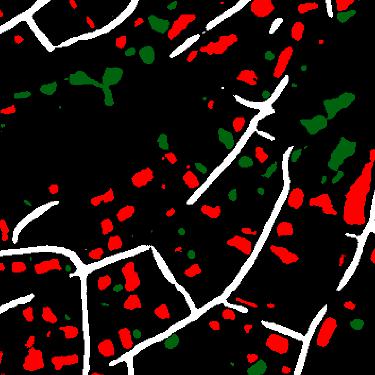}}} &
\raisebox{-.5\height}{\frame{\includegraphics[width=0.135\linewidth]{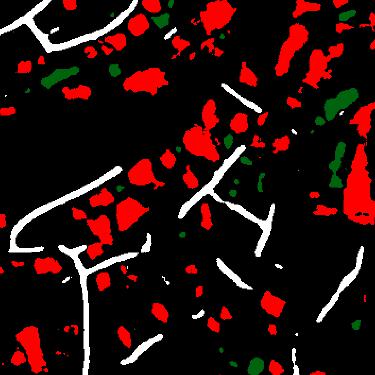}}} &
\raisebox{-.5\height}{\frame{\includegraphics[width=0.135\linewidth]{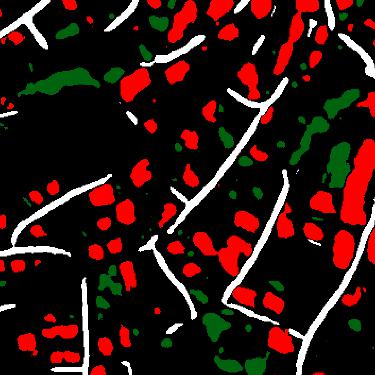}}} & 
\raisebox{-.5\height}{\frame{\includegraphics[width=0.135\linewidth]{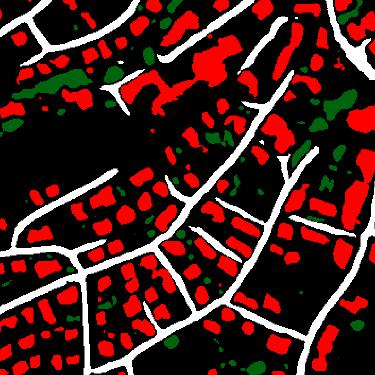}}}\\
\\[-3.5mm]

\rotatebox[origin=c]{90}{Leibnitz}&
\raisebox{-.5\height}{\frame{\includegraphics[width=0.135\linewidth]{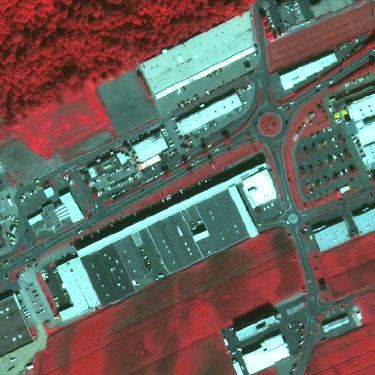}}} &
\raisebox{-.5\height}{\frame{\includegraphics[width=0.135\linewidth]{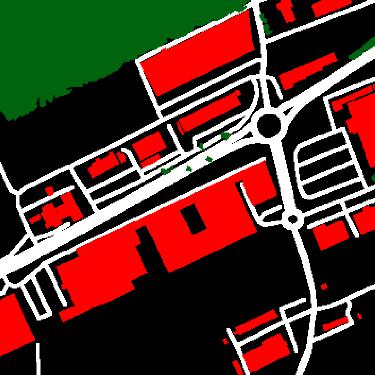}}} &
\raisebox{-.5\height}{\frame{\includegraphics[width=0.135\linewidth]{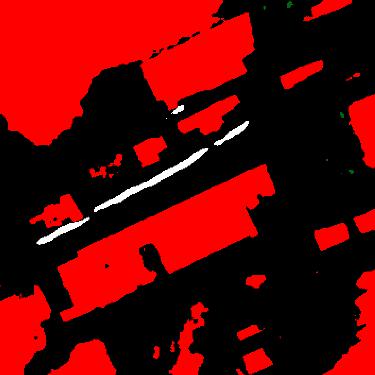}}} &
\raisebox{-.5\height}{\frame{\includegraphics[width=0.135\linewidth]{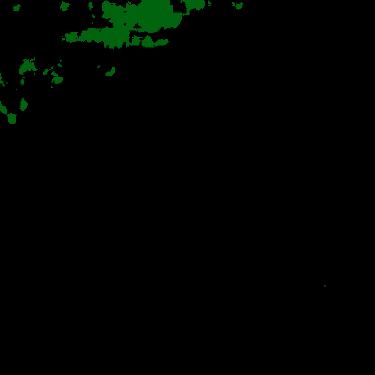}}} &
\raisebox{-.5\height}{\frame{\includegraphics[width=0.135\linewidth]{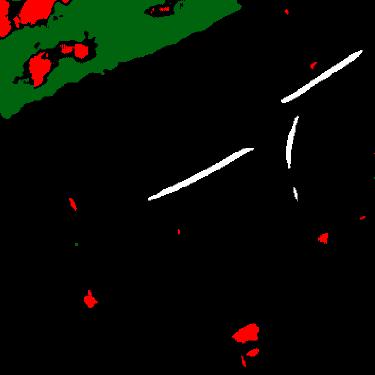}}} &
\raisebox{-.5\height}{\frame{\includegraphics[width=0.135\linewidth]{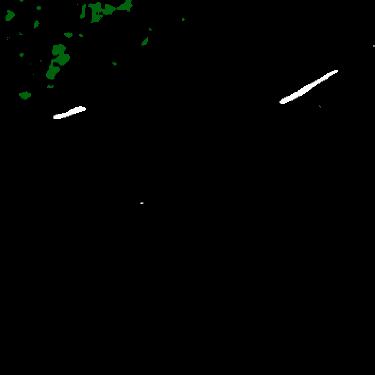}}} & 
\raisebox{-.5\height}{\frame{\includegraphics[width=0.135\linewidth]{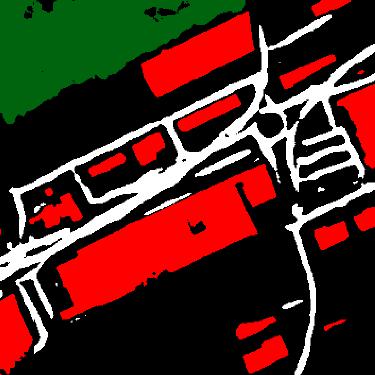}}}\\
\end{tabular}
\caption{Target images used in the third experiment, their ground-truth, and the predictions. Red, white, and green pixels represent building, road, and tree classes, respectively. The pixels in black do not belong to any class.}
\label{fig:exp3_preds}
\end{figure*}

\subsection{Life-long Domain Adaptation}
In the third experiment, we assume that in addition to the ones in the second experiment, we receive new annotated source images from B{\'e}ziers, Salzburg~Stadt, Albi, and unlabeled target data from Leibnitz. The image from Leibnitz consists of near-infrared, red, and green channels, whereas the other nine cities are composed of red, green, and blue spectral bands. This experimental setup has two goals. Firstly, we wish to verify whether our method can perform style transfer between ten images having different data distributions even when the type of spectral bands are different. Secondly, we want to observe the relevance of our solution for the life-long adaptation setting, where new source and target images are added. In this experimental setup, we have seven source and three target cities in total.

We first generate unique style codes for the new cities and train our style transfer method as described in Sec.~\ref{sec:daugnet} for 50 epochs. As in the first two experiments, we use Adam optimizer with the initial learning rate of 0.0001. We reduce the learning rate after the 25$^{th}$ epoch via Eq.~\ref{eq:lr}. Before training DAugNet, we initialize the parameters of U-net in DAugNet with the pre-trained weights from the second experiment. We then fine-tune DAugNet for 100 epochs. We compare DAugNet with the same data standardization approaches as the previous experiment. For each compared standardization method, we fine-tune the U-net trained in the second experiment for 100 epochs.

As confirmed by Table~\ref{table:iou_exp3}, because of the new three source cities, the IoU scores of all the methods for Bad~Ischl and Vaduz are higher than in the second experiment. For example, the performance of U-net especially for tree class significantly improves. It is probably because trees in the new cities are more representative for the target cities. For instance, as can be seen in Fig.~\ref{fig:exp3_md_style_transfer}, the data distributions of trees in Salzburg~stadt and Vaduz, and Albi and Bad~Ischl seem close. Naive U-net slightly better detects trees in Vaduz than DAugNet. This slight performance difference probably stems from some artifacts added by our GAN architecture in the process of generating fake cities. However, DAugNet outperforms the compared approaches for all the other classes. When it comes to segmenting Leibnitz, the performances of the compared methods are extremely poor, mainly because the spectral bands of this image do not comprise red, green, and blue channels. On the other hand, as shown in Fig.~\ref{fig:exp3_md_style_transfer}, our style transfer method can easily convert red, green, blue images to near infrared, red, green ones. As a consequence, DAugNet significantly outperforms the others on Leibnitz, since it learns from data consisting of both red, green, blue and near infrared, red, green bands. Fig.~\ref{fig:exp3_preds} depicts a close-up from each city used in this experiment, its ground-truth, and the predictions.

\begin{figure*}
\centering
\subfloat[]{\includegraphics[width=0.33\linewidth]{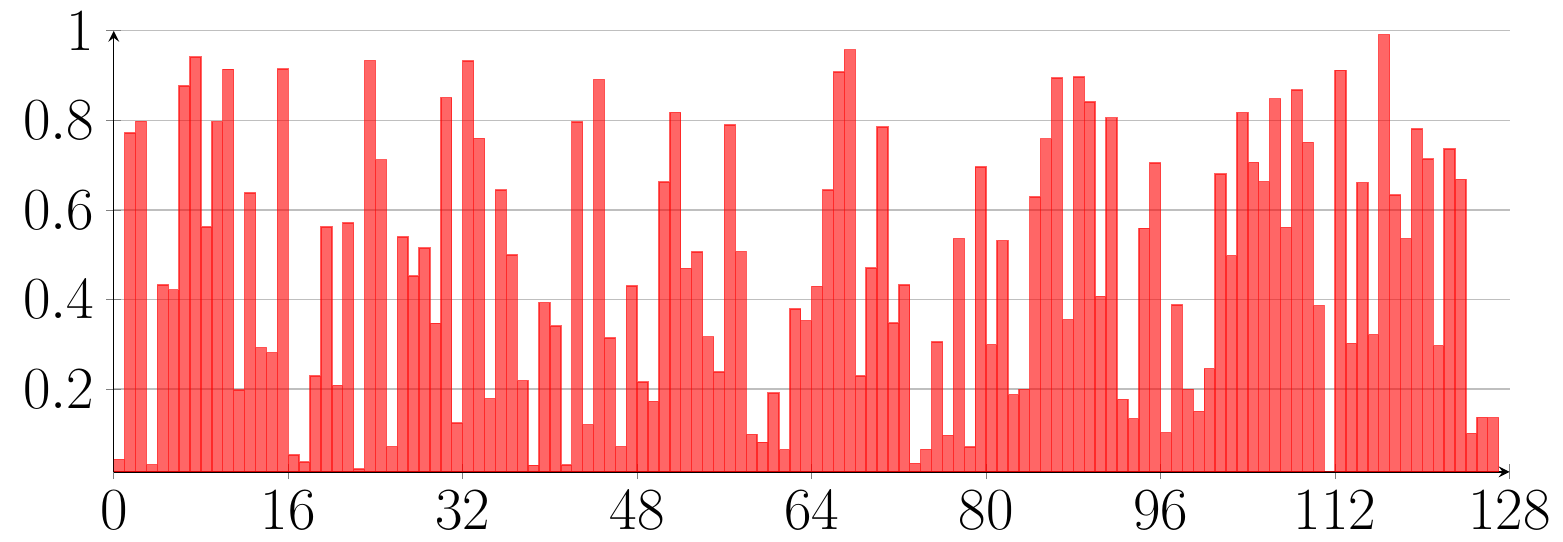}}
\hfill
\subfloat[]{\includegraphics[width=0.33\linewidth]{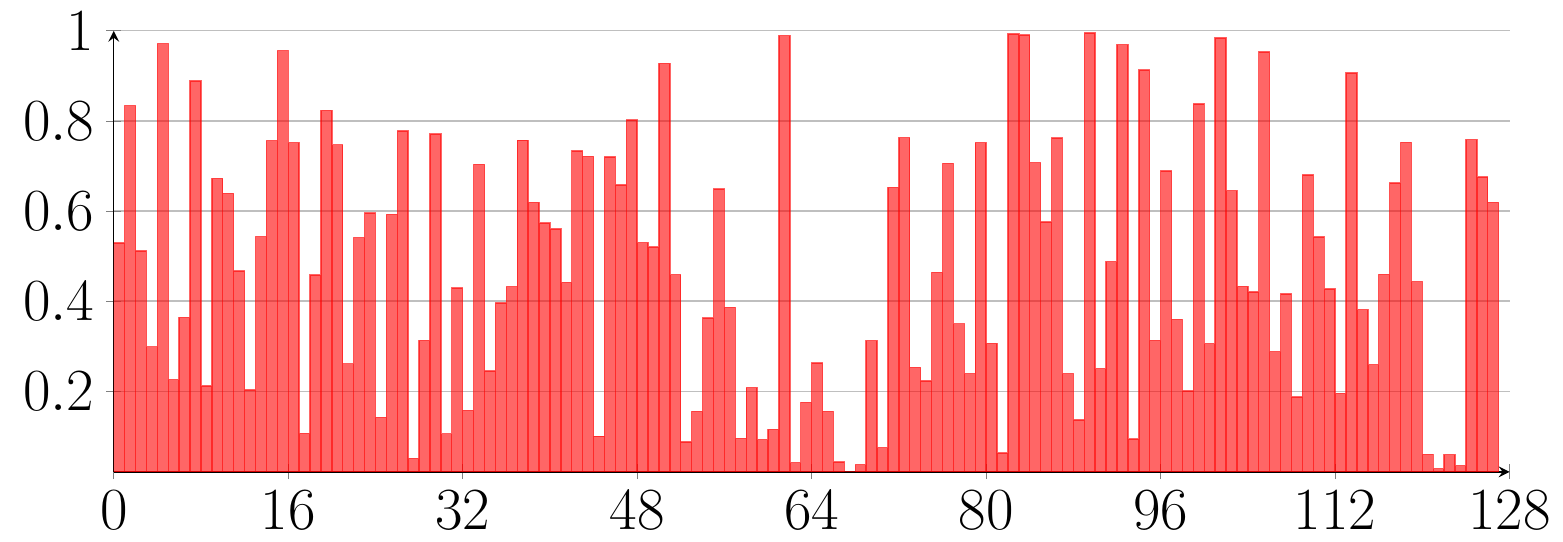}}
\hfill
\subfloat[]{\includegraphics[width=0.33\linewidth]{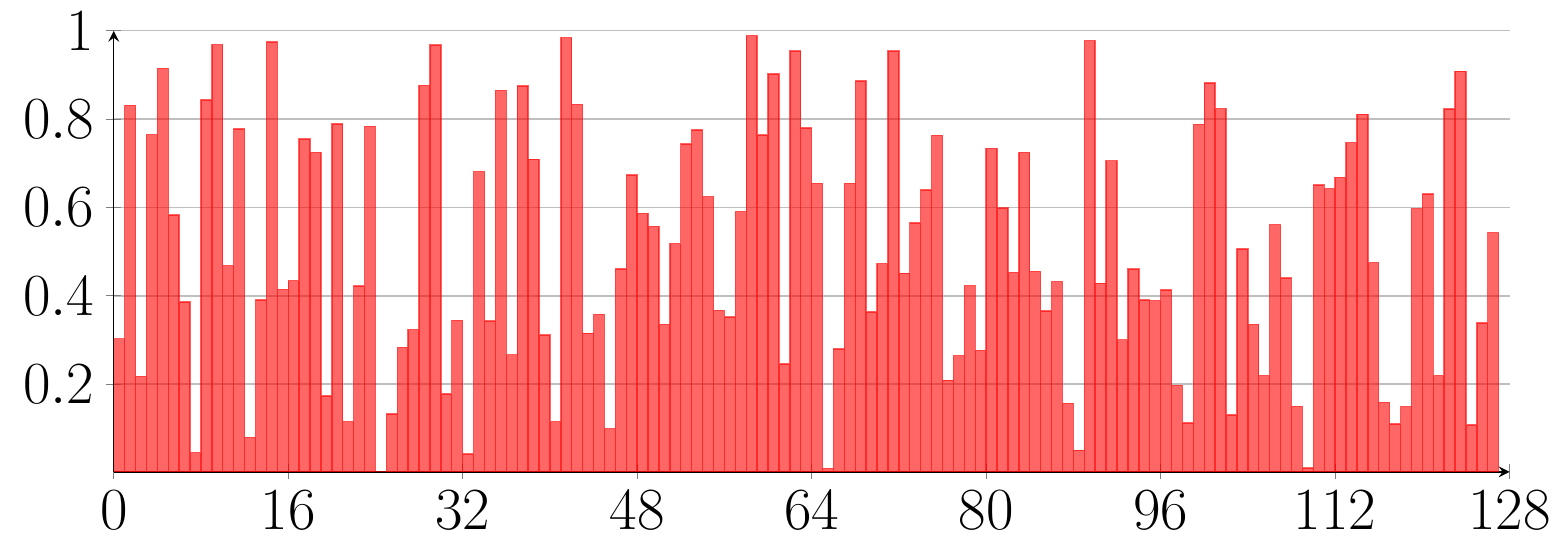}}
\hfill
\subfloat[]{\includegraphics[width=0.33\linewidth]{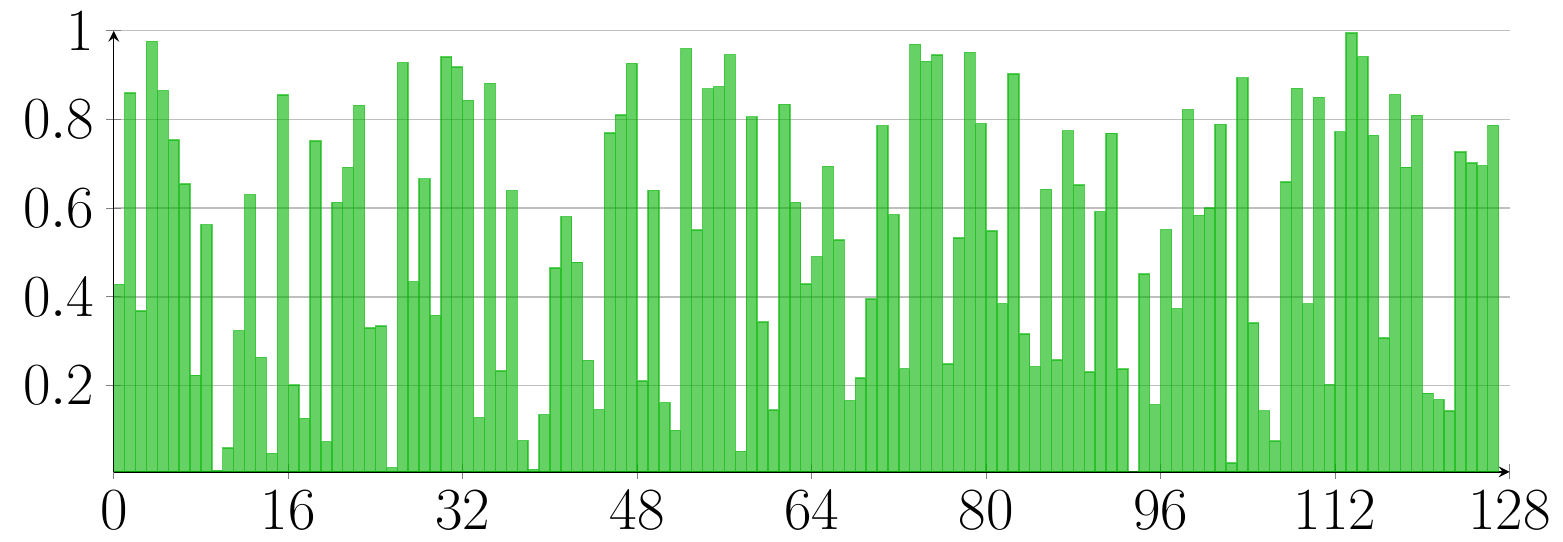}}
\hfill
\subfloat[]{\includegraphics[width=0.33\linewidth]{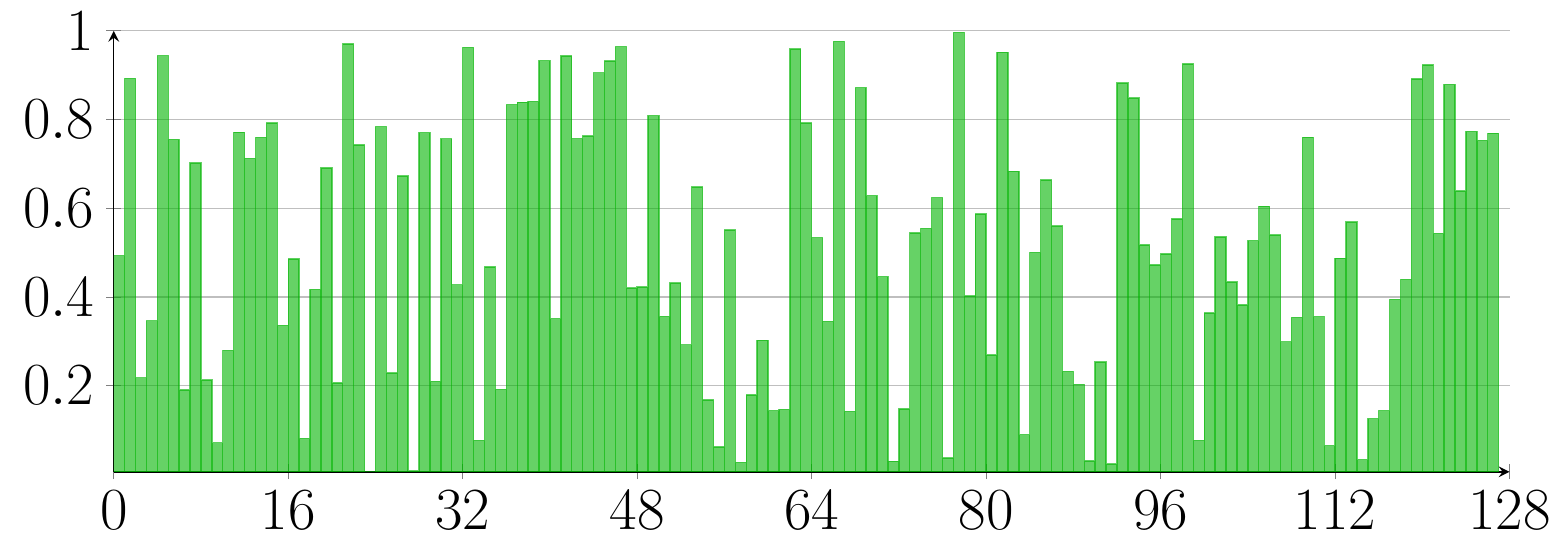}}
\hfill
\subfloat[]{\includegraphics[width=0.33\linewidth]{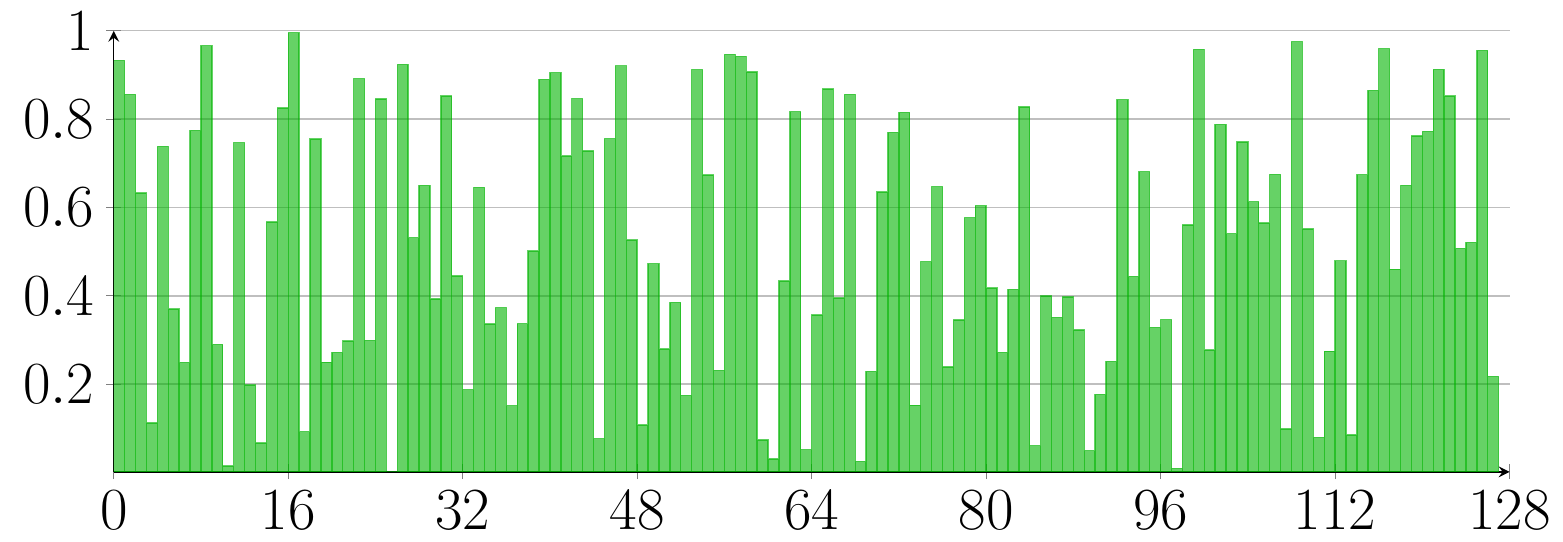}}
\caption{Random style codes of Villach in the second experimental setup and in the second ablation study. (a-d) $\gamma$ and $\beta$ values of Villach in the second experiment. (b-e) $\gamma$ and $\beta$ values of Villach in the first run of the second ablation study. (c-f) $\gamma$ and $\beta$ values of Villach in the second run of the same study.}
\label{fig:random_gamma_beta}
\end{figure*}

\begin{table}
\centering
\caption{Quantitative results for the first ablation study.}
\label{table:ablation_1}
\begin{tabular}{||c|c|cccc||}
\hline
\textbf{City} & \textbf{Result~type} & \textbf{Building} & \textbf{Road} & \textbf{Tree} & \textbf{Overall} \\
\hline  
\multirow{3}{*}{Bad~Ischl}
& IoUs for run 2 & 59.91 & 50.00 & 82.70 & 64.20 \\
& IoUs for run 3 & 60.50 & 50.44 & 83.64 & 64.86 \\
& standard dev.  &  0.34 &  0.58 &  0.40 &  0.35 \\ 
\hline
\hline
\multirow{3}{*}{Vaduz}
& IoUs for run 2 & 51.65 & 35.42 & 62.56 & 49.88 \\ 
& IoUs for run 3 & 51.72 & 32.76 & 66.98 & 50.49 \\
& standard dev.  &  0.07 &  1.10 &  2.96 &  0.89 \\
\hline
\end{tabular}
\end{table}

\begin{table}
\centering
\caption{Quantitative results for the second ablation study.}
\label{table:ablation_2}
\begin{tabular}{||c|c|cccc||}
\hline
\textbf{City} & \textbf{Result~type} & \textbf{Building} & \textbf{Road} & \textbf{Tree} & \textbf{Overall} \\
\hline  
\multirow{3}{*}{Bad~Ischl}
& IoUs for run 2 & 59.10 & 49.45 & 81.71 & 63.42 \\
& IoUs for run 3 & 59.95 & 52.27 & 83.16 & 65.12 \\
& standard dev.  &  0.36 &  1.43 &  0.74 &  0.70 \\
\hline
\hline
\multirow{3}{*}{Vaduz}
& IoUs for run 2 & 50.07 & 37.31 & 64.80 & 50.73 \\ 
& IoUs for run 3 & 46.03 & 34.14 & 65.86 & 48.68 \\
& standard dev.  &  2.43 &  1.43 &  2.12 &  1.36 \\
\hline
\end{tabular}
\end{table}

\begin{figure}
\centering
\subfloat[]{\includegraphics[width=0.33\linewidth]{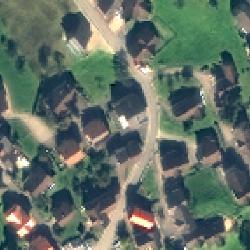}}
\hfill
\subfloat[]{\includegraphics[width=0.33\linewidth]{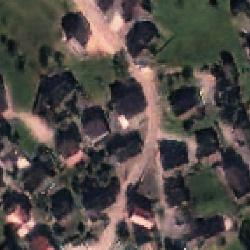}}
\hfill
\subfloat[]{\includegraphics[width=0.33\linewidth]{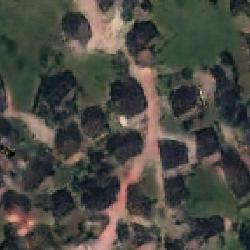}}
\caption{Real Vaduz and Fake Vaduz in Bad~Ischl style. (a) Vaduz, fake Vaduz by our approach with (b) edge loss and (c) without edge loss.}
\label{fig:edge_loss_effect}
\end{figure}

\begin{table}
\centering
\caption{Drop in IoU scores when edge loss is deactivated.}
\label{table:ablation_3}
\begin{tabular}{||c|cccc||}
\hline
\textbf{City} & \textbf{Building} & \textbf{Road} & \textbf{Tree} & \textbf{Overall} \\
\hline  
Bad~Ischl & 2.04 & 2.25 &  2.59 & 2.30 \\
Vaduz     & 5.88 & 2.00 & 14.81 & 7.56 \\
\hline
\end{tabular}
\end{table}

\begin{figure*}
\centering
\begin{tabular}{p{0.1em}c}
& \large{\textbf{STYLE}} \\
\rotatebox{90}{\large{\textbf{CONTENT}}}  & \raisebox{-.5\height}{\includegraphics[width=0.965\linewidth]{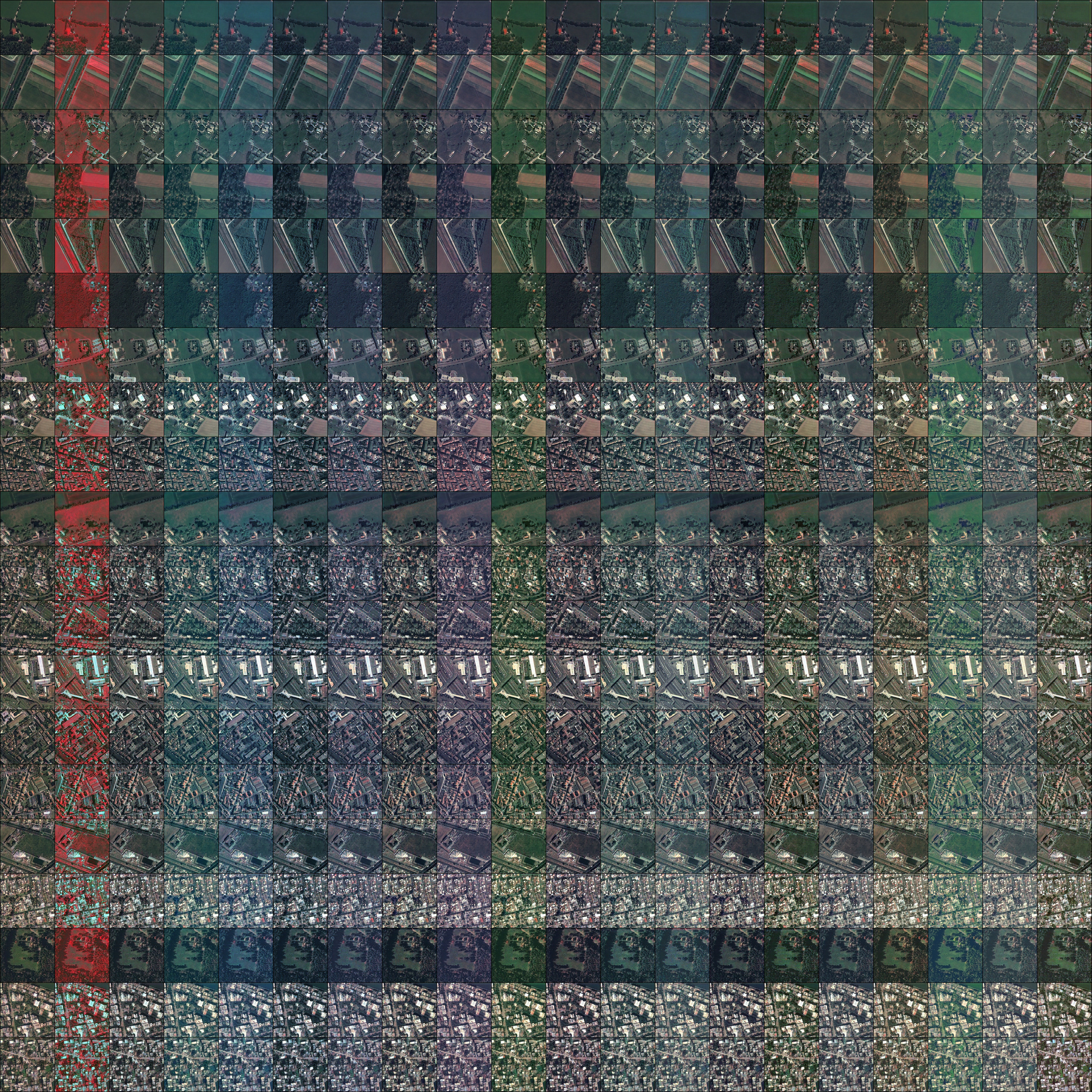}}
\end{tabular}
\caption{Style transfer between 20 satellite images, each of which has been collected from a different city. Rows and columns of this matrix correspond to content and style, respectively. The cells on the diagonal from top-left to bottom-right represent real images. Each of the rest of the cells shows a fake city with the style of another one.}
\label{fig:ablation_5}
\end{figure*}

\subsection{Ablation Studies}
\paragraph{Effect of random diversification} As explained earlier, the data augmentor in DAugNet randomly diversifies the training batch before passing it to the classifier. In the first ablation study, we analyze the robustness of DAugNet to random diversification. We train DAugNet two more times from scratch on the source cities in the second experimental setup and segment Bad~Ischl and Vaduz. Note that we do not train the initial multi-domain style transfer part of our pipeline again to evaluate the effect of only random diversification. Table~\ref{table:ablation_1} reports IoU scores of the two runs. In the table, we also indicate the standard deviation of IoUs in the second experiment and IoUs of these two runs. Because the standard deviations are quite small, we conclude that our framework is robust to random diversification.

\paragraph{Effect of random style codes and diversification} We repeat the whole training pipeline of our approach in the second experiment two times. Fig.~\ref{fig:random_gamma_beta} depicts the randomly initialized 128 $\gamma$ and $\beta$ values for Villach in the second experiment and in the two runs of this ablation study. In Table~\ref{table:ablation_2}, we report IoU scores of these two runs and the standard deviations. Although the style codes (i.e., $\gamma$ and $\beta$) of the cities are different in each run, the standard deviations for IoUs are very small. In conclusion, our solution is robust to random style codes and random diversification.  

\paragraph{Effect of edge loss} We deactivate the edge loss by setting $\lambda_4$ in Eq.~\ref{eq:g_loss} to zero and repeat the second experiment to evaluate the effect of the edge loss. As can be seen in Fig.~\ref{fig:edge_loss_effect}, when $\lambda_4$ is zero, our style transfer approach generates slightly blurred fake images, which causes DAugNet to exhibit a worse performance. The drop in IoU scores is reported in Table~\ref{table:ablation_3}. Note that the other terms in Eq.~\ref{eq:g_loss} are mandatory; therefore, $\lambda_1$, $\lambda_2$, and $\lambda_3$ cannot be set to zero.

\paragraph{U-net vs. DAugNet training time comparison} Because the data augmentor needs to diversify the batch in each training iteration, the training time of DAugNet is longer than that of U-net. Table~\ref{table:ablation_4} reports the time required to train DAugNet and U-net on the source data in the second and in the third experiments on an Nvidia GeForce GTX 1080 Ti GPU. Due to its simple architecture, the data augmentor prolongs training time of the classifier for only 18 minutes in the second experiment, and for 28 minutes in the third experiment. Table~\ref{table:tr_time_md_style_transfer} reports the time needed to train the multi-domain style transfer part of our framework in the second and in the third experiments.

\paragraph{Scalability of the proposed method} As mentioned in Sec.~\ref{sec:introduction}, significant technological advancements in satellite sensors and a large number of satellite missions have made massive volume of remote sensing data available. Therefore, it is of paramount importance to propose highly scalable methods that can efficiently process a lot of satellite images. To demonstrate that our method is capable of dealing with many satellite images, we perform style transfer between twenty images collected from different cities in Europe. Spectral bands of one image consists of near-infrared, red, green bands, whereas the other images comprise red, green, and blue channels. As depicted in Fig.~\ref{fig:ablation_5}, our style transfer method successfully stylizes each images like any other. This ablation study proves that our method is suitable for large-scale segmentation problems.

\begin{table}
\centering
\caption{DAugNet vs. U-net training time comparison.}
\label{table:ablation_4}
\begin{tabular}{||c|c|c@{\hspace{0.8em}}|c@{\hspace{0.8em}}||}
\hline
\multirow{2}{*}{\textbf{Training Data ($\#$ cities)}}   & \multirow{2}{*}{\textbf{$\#$ patches}} 
& \multicolumn{2}{c||}{\textbf{Tr.~time (h:m:s)}} \\
\cline{3-4}
& & \textbf{U-net} & \textbf{DAugNet}\\
\hline
Source data in Exp.~2 (4) & 5676 & 1:47:21 & 2:05:25 \\
Source data in Exp.~3 (7) & 8992 & 2:50:29 & 3:18:05 \\
\hline
\end{tabular}
\end{table}

\begin{table}
\centering
\caption{Training time of multi-domain style transfer.}
\label{table:tr_time_md_style_transfer}
\begin{tabular}{||c|c|c|c||}
\hline
\textbf{Tr. Data ($\#$ cities)} & \textbf{$\#$ patches} & \textbf{$\#$ epochs}  & \textbf{Tr.~time (h:m:s)} \\
\hline
Cities in Exp.~2 (6)  &  7745 & 200 & 2:25:47 \\
Cities in Exp.~3 (10) & 11633 &  50 & 0:34:57 \\
\hline
\end{tabular}
\end{table}

\section{Concluding Remarks}
The satellite missions launched in the last decade have enabled us to collect huge volume of remote sensing data over the whole earth every day. Because of various atmospheric conditions, sensor characteristics, and differences in acquisition, remote sensing images collected from separate geographic locations in distinct times tend to have largely different data distributions. The data distribution difference between source and target images prevents the machine learning models from generating precise maps. Moreover, scarcity of annotated data motivates us to propose methods that are robust to such distribution difference. In addition, it is of paramount importance to introduce new approaches that can adapt to continually growing data.

In this context, we presented a new approach for multi-source, multi-target, and life-long domain adaptation problem. In the pre-processing stage of our method, we learn how to perform style transfer between multiple source and target domains using only one encoder, one decoder, and one discriminator. A subset of the networks used in this stage constitutes a data augmentor. We then train the proposed DAugNet comprising the data augmentor and a classifier. In each training iteration, the data augmentor randomly diversifies the training batch before passing it to the classifier. As a consequence, the classifier is more robust to the data distribution difference between the images, since it learns from the data that are representative for all source and target cities. 

The state-of-the-art methods in the literature either aim at solving single-source and single target adaptation problem or use multiple networks to deal with multi-source domain adaptation problem. However, because the number of networks in multi-domain style transfer stage (pre-processing step) of our method is constant regardless of the number of cities, our approach enabled us to perform adaptation between many cities. For the same reason, we could extend our solution to life-long adaptation setting. In three extensive experiments, we verified effectiveness of our solution. Moreover, in five ablation studies, we analyzed the properties of our approach in great detail. One limitation of our approach is that it cannot be trained end-to-end. The pre-processing step and DAugNet need to be trained consecutively. In addition, the training time of DAugNet is slightly longer than that of U-net. However, as confirmed by Table~\ref{table:ablation_4}, the difference is kept small due to the simple architectural design of the data augmentor.

As future work, we plan to investigate between how many cities our approach can perform style transfer. We will also observe whether our solution can be used for domain adaptation of other types of remote sensing data such as aerial and Sentinel images.

\bibliographystyle{IEEEtran}
\bibliography{refs}

\begin{IEEEbiography}[{\includegraphics[width=1in,
height=1.25in,clip,keepaspectratio]{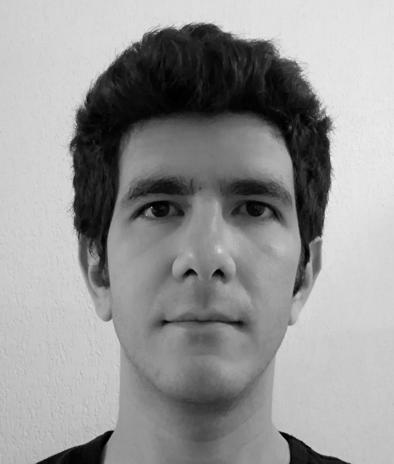}}]{Onur Tasar} (Student Member, IEEE) received the B.S. degree in computer engineering department from Hacettepe University, Ankara, Turkey in 2014, and the M.S. degree in computer engineering department from Bilkent University, Ankara, Turkey in 2017. He is currently working towards his Ph.D. at Inria Sophia Antipolis-M\'editerran\'ee within the Titane project-team, Valbonne, France.

His research interests include computer vision, machine learning, and computational geometry with applications to remote sensing.
\end{IEEEbiography}

\vspace{-2mm}

\begin{IEEEbiography}[{\includegraphics[width=1in,
height=1.25in,clip,keepaspectratio]{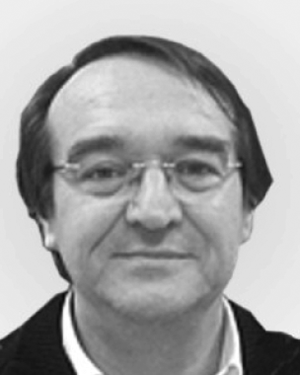}}]{Alain Giros} received the physics engineer degree from {\'E}cole Nationale Sup{\'e}rieure de Physique de Marseille, Marseille, France, in 1982, with a specialization in optics and image processing. He has since been with the Centre National d'{\'E}tudes Spatiales (CNES), Toulouse, France.

He began his carrier in the SPOT 1 project team where he was responsible of the whole image processing software. Then, he was responsible for the development of several image processing systems, including DALI (the worldwide catalog of SPOT images), several image quality assessment systems for SPOT, and the Vegetation User Services Facility. Interested in automated image registration since the mid-1990s, he has been promoting this research subject at CNES. He is now in charge of the development of image processing techniques mainly in the field of image time series processing, automated registration, and image information mining.
\end{IEEEbiography}

\vspace{-2mm}

\begin{IEEEbiography}[{\includegraphics[width=1in,
height=1.25in,clip,keepaspectratio]{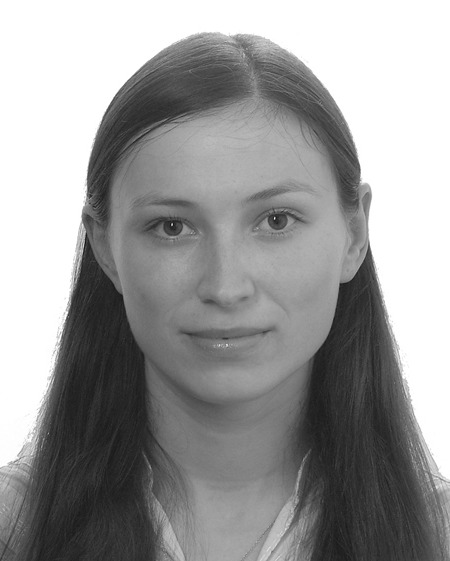}}]{Yuliya Tarabalka} (Senior Member, IEEE) received the B.S. degree in computer science from Ternopil Ivan Pul'uj State Technical University, Ukraine, in 2005 and the M.Sc. degree in signal and image processing from the Grenoble Institute of Technology (INPG), France, in 2007. She received a joint Ph.D. degree in signal and image processing from INPG and in electrical engineering from the University of Iceland, in 2010.

From July 2007 to January 2008, she was a researcher with the Norwegian Defence Research Establishment, Norway. From September 2010 to December 2011, she was a postdoctoral research fellow with the Computational and Information Sciences and Technology Office, NASA Goddard Space Flight Center, Greenbelt, MD. From January to August 2012 she was a postdoctoral research fellow with the French Space Agency (CNES) and Inria Sophia Antipolis-M\'editerran\'ee, France. From 2012 to 2019 she was a researcher with the Titane project-team of Inria Sophia Antipolis-M\'editerran\'ee. She is currently the research director of LuxCarta Technology. Her research interests are in the areas of image processing, pattern recognition and development of efficient algorithms. She is Member of the IEEE Society.
\end{IEEEbiography} 

\vspace{-2mm}

\begin{IEEEbiography}[{\includegraphics[width=1in,
height=1.25in,clip,keepaspectratio]{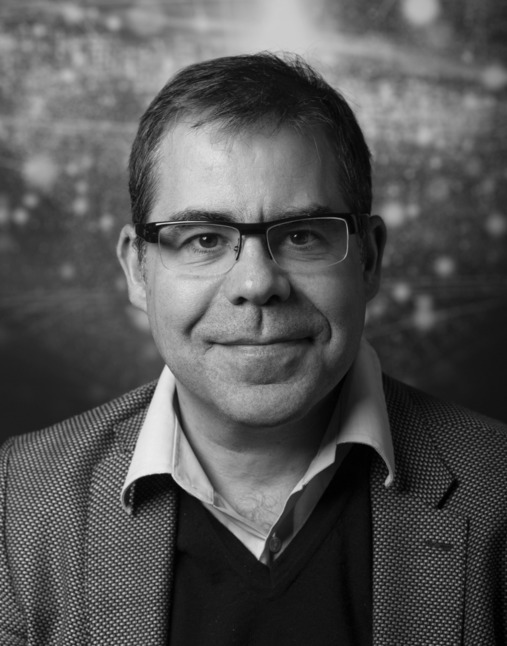}}]{Pierre Alliez}
is Senior Researcher, Team Leader and Head of Science at Inria Sophia-Antipolis - M\'editerran\'ee. He has authored several scientific publications and book chapters on topics commonly referred to as geometry processing, and on analysis of remote sensing images. He was awarded in 2005 the EUROGRAPHICS young researcher award for his contributions to computer graphics and geometry processing. He was co-chair of the Symposium on Geometry Processing in 2008, Pacific Graphics in 2010, Geometric Modeling and Processing in 2014 and EUROGRAPHICS annual conference in 2019. He was awarded in 2011-2015 a Starting Grant, and in 2017-2018 a Proof of Concept Grant, from the European Research Council on Robust Geometry Processing.
\end{IEEEbiography}

\vspace{-2mm}

\begin{IEEEbiography}[{\includegraphics[width=1in,
height=1.25in,clip,keepaspectratio]{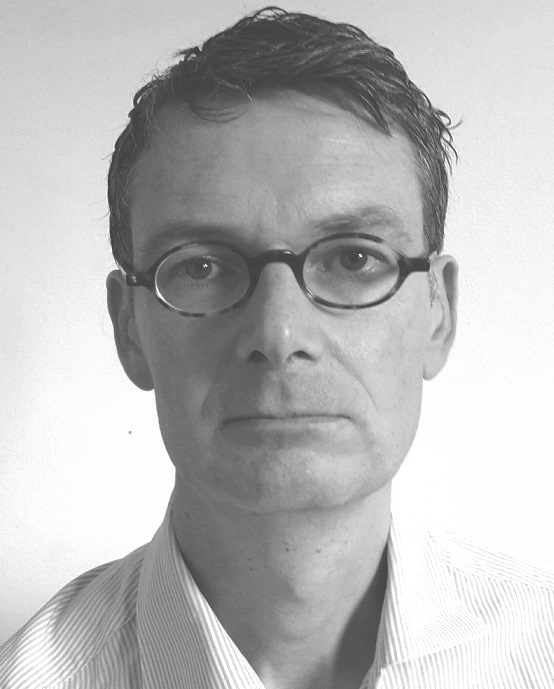}}]{S{\'e}bastien Clerc}received the M.Sc. degree from the Ecole Polytechnique, Palaiseau, France in 1994, and the Ph.D. degree in numerical analysis from Universit{\'e} Pierre et Marie Curie (also known as Paris 6), Paris, France, in 1997. He was a Research Engineer with the Commissariat {\`a} l’Energie Atomique, Saclay, before joining Thales Alenia Space, Cannes, where he worked as an AOCS Engineer and Systems Engineer. He joined ACRI-ST, Sophia Antipolis, France, in 2014, and has been working there. His main activity is the technical management of the Sentinel-2 Mission Performance Center, a team of experts and operators monitoring and improving the quality of the mission products. He was a Panelist for the first ECCOE Workshop in Sioux Falls in 2017.
\end{IEEEbiography}

\end{document}